\documentclass[10pt,twocolumn,letterpaper]{article}
\usepackage[pagenumbers]{cvpr}

\usepackage{balance}

\usepackage{algorithm}
\usepackage{algpseudocode}

\usepackage{array}

\definecolor{cvprblue}{rgb}{0.21,0.49,0.74}
\usepackage[pagebackref,breaklinks,colorlinks,allcolors=cvprblue]{hyperref}
\def\confName{CVPR}
\def\confYear{2027}
\title{RetroNav: Scene-Grounded Retrospection and Retry for Self-Evolving Lifelong Navigation}
\author{
Diantao Tu\textsuperscript{1}\qquad
Yangdong Liu\textsuperscript{1}\qquad
Niannian Dai\textsuperscript{1,3}\qquad
Mengqi Rong\textsuperscript{1}\thanks{Corresponding authors: Mengqi Rong (mengqi.rong@nlpr.ia.ac.cn) and Shuhan Shen (shshen@nlpr.ia.ac.cn).}\qquad
Shuhan Shen\textsuperscript{1,2}\footnotemark[1]\\
\normalsize\textsuperscript{1}Institute of Automation, Chinese Academy of Sciences\\
\normalsize\textsuperscript{2}School of Artificial Intelligence, University of Chinese Academy of Sciences\\
\normalsize\textsuperscript{3}School of Advanced Interdisciplinary Sciences, University of Chinese Academy of Sciences\\
{\ttfamily\small\{diantao.tu, yangdong.liu\}@ia.ac.cn, dainiannian22@mails.ucas.ac.cn}\\
{\ttfamily\small\{mengqi.rong, shshen\}@nlpr.ia.ac.cn}
}
\begin{document}
\maketitle
\begin{abstract}
Existing robot navigation methods primarily focus on isolated tasks, overlooking the need to accumulate knowledge and experience across repeated navigation episodes. 
Although recent lifelong navigation methods have explored cross-task knowledge accumulation, they mainly focus on scene knowledge while neglecting historical navigation experience and its connection to scene evidence, limiting the ability to reflect on past mistakes and improve future decisions.
To address this limitation, this paper presents \textbf{RetroNav}, a training-free self-evolving navigation framework that integrates navigation, memory accumulation, and failure reflection. Specifically, we develop a unified memory system that connects geometric and semantic scene knowledge with historical navigation experience, distinguishing objective observations from task-dependent interpretations. 
After each navigation attempt, an independent evaluator determines whether a retry is necessary. If so, a Retrospection Agent revisits historical decisions using accumulated scene knowledge and navigation records, generating an evidence-grounded retry brief to guide the next attempt. 
RetroNav achieves state-of-the-art performance across five navigation benchmarks.
Our method will be open-sourced at {\hypersetup{urlcolor=black}\url{https://github.com/3dv-casia/RetroNav}}.
\end{abstract}

\section{Introduction}
\label{sec:intro}

\begin{figure*}[t]
    \centering
    \includegraphics[width=\textwidth]{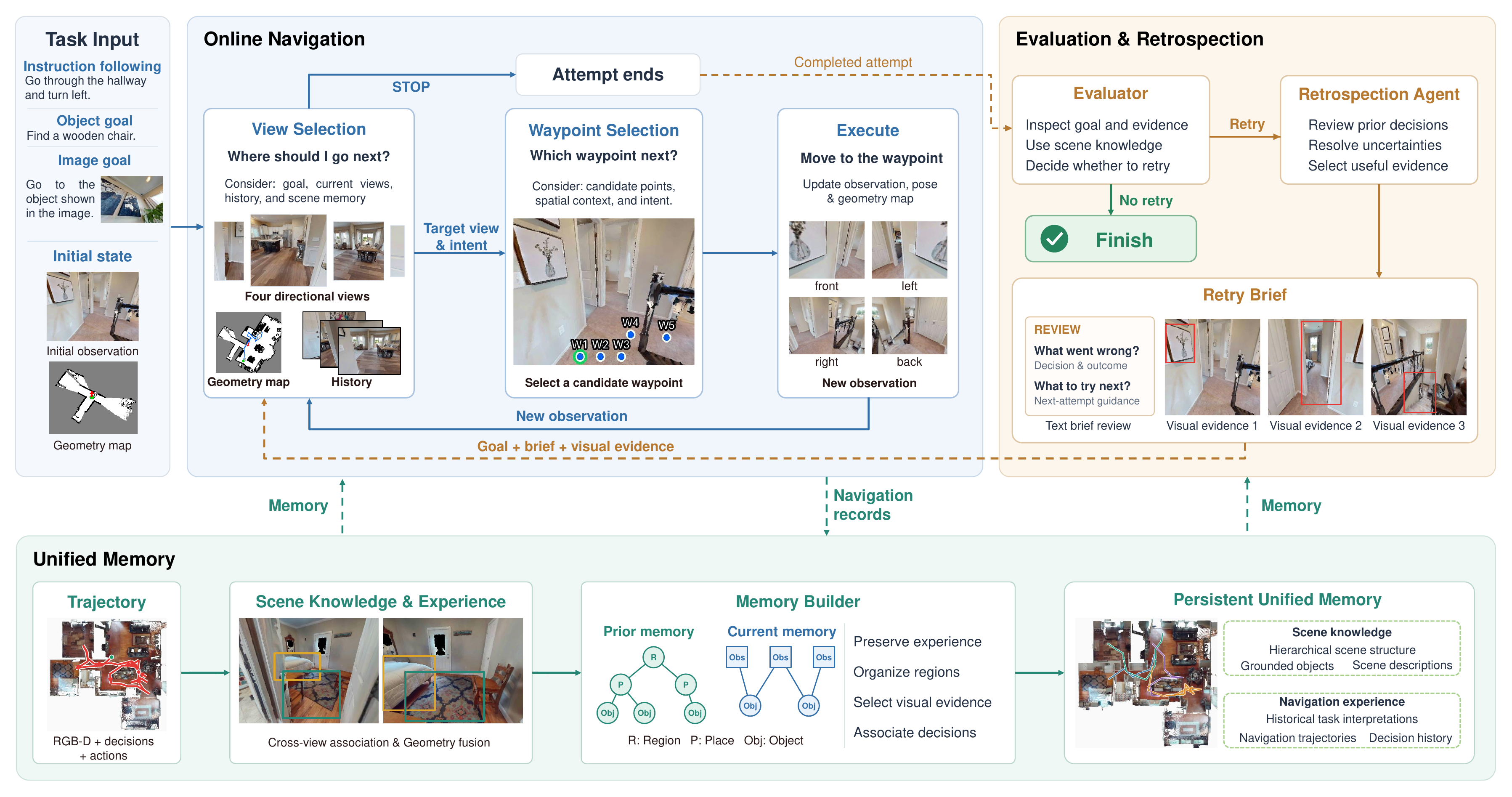}
    \caption{Overview of the proposed RetroNav framework. The navigator iteratively selects navigation directions and waypoints, while a unified memory system accumulates scene knowledge and historical navigation experience. After each navigation attempt, an independent Evaluator determines whether a retry is necessary. When triggered, a Retrospection Agent revisits historical decisions using accumulated scene knowledge and visual evidence, generating an evidence-grounded Retry Brief to guide the next attempt.}
    \label{fig:pipeline}
\end{figure*}

Most existing robot navigation research focuses on isolated navigation tasks. In practical applications, however, robots often operate in the same environment over long periods, repeatedly performing navigation tasks and encountering familiar places and objects. 
Each navigation episode, whether successful or not, may provide valuable environmental information for future tasks. Therefore, navigation systems should not only complete individual tasks but also accumulate knowledge and experience to reduce repeated exploration and improve subsequent decisions.

Recent studies have begun to investigate lifelong Navigation, exploring how robots can leverage historical information across tasks~\cite{krantz2023ivln,khanna2024goat,yang2025mem,ge2026evomemnav,zhao2026navharness}. 
Existing methods primarily focus on accumulating scene knowledge, such as geometric structures, semantic information, and visual observations, while largely overlooking navigation experience, including how tasks were interpreted and decisions were made. 
However, navigation failures may arise not only from insufficient environmental knowledge but also from incorrect interpretations or decisions. 
For example, a robot may have already observed the correct target location but choose a wrong direction due to misinterpreting the instruction. Simply retaining scene knowledge cannot reveal such decision errors, while reusing previous decisions may repeat them. Therefore, effective lifelong navigation requires associating historical decisions with their underlying scene evidence, enabling navigation experience to be revisited and reused.

We argue that effective lifelong navigation requires integrating scene knowledge with historical navigation experience. 
Scene knowledge captures the geometric and semantic structure of the environment, providing reusable information for future navigation. Navigation experience records how previous tasks were interpreted and executed, enabling robots to revisit historical decisions and reflect on potential errors. 
By associating navigation experience with scene evidence, robots can continuously improve their navigation behavior through knowledge accumulation and failure reflection.

To this end, we propose RetroNav, a training-free self-evolving navigation framework that integrates navigation execution, memory accumulation, and failure reflection into a unified system, as illustrated in Fig.~\ref{fig:pipeline}. 
At each navigation step, the View Selection Agent interprets navigation instructions and selects a direction, while the Waypoint Selection Agent determines a suitable waypoint from geometrically feasible candidates. 
Meanwhile, a Memory Builder accumulates scene knowledge from visual and geometric observations, while preserving navigation experience that captures historical decisions, their reasoning, and supporting scene evidence. 
By linking scene knowledge with navigation experience, the unified memory supports both future navigation and retrospective reasoning. After each navigation attempt, an independent Evaluator assesses the navigation outcome and determines whether a retry is necessary. If so, a Retrospection Agent re-examines previous decisions using accumulated scene knowledge and navigation experience, generating an evidence-grounded Retry Brief to guide the next attempt.

Our contributions are threefold:
\begin{itemize}
    \item We propose RetroNav, a training-free self-evolving navigation framework that integrates navigation, memory accumulation, and failure reflection for lifelong navigation.
    \item We develop a unified memory system that associates scene knowledge with historical navigation experience, enabling knowledge reuse and evidence-grounded reflection.
    \item We introduce an evaluation-driven retry mechanism that reflects on historical decisions using accumulated scene knowledge and navigation experience to guide subsequent attempts.
\end{itemize}

\section{Related Work}
\label{sec:related}

\textbf{Lifelong navigation in persistent environments.}
Lifelong navigation studies ordered tasks executed in a persistent scene, where later tasks can reuse information collected during earlier episodes. While R2R~\cite{anderson2018vln} and RxR~\cite{ku2020rxr} established visually grounded instruction following, IVLN/IR2R~\cite{krantz2023ivln} explicitly evaluates navigation over ordered episodes. 
GOAT-Bench~\cite{khanna2024goat} extends this setting to multimodal goals. It also studies explicit and implicit scene memory. OVAL~\cite{pei2026oval} uses past observations to find successive open-vocabulary object goals. It guides exploration using semantic cues, distance, and previous visits.
EvoMemNav~\cite{ge2026evomemnav} reflects after each subtask to update its environmental priors. NavHarness~\cite{zhao2026navharness} retains maps and task experience for later tasks and recovery attempts.
These methods improve navigation through persistent scene knowledge, continual policy learning, and experience reuse. RetroNav separates environmental evidence from task-dependent interpretations and decisions, while explicitly linking them for later inspection. This design supports evidence-based failure diagnosis and targeted retries through independent post-attempt evaluation and a structured Retry Brief.

\textbf{Scene memory construction.}
Scene-memory methods differ in the representations they retain and the mechanisms used to consolidate them. 
UniGoal~\cite{yin2025unigoal} incrementally constructs a scene graph of objects and their spatial or semantic relations, and matches a goal graph against it to guide navigation, while structured representations connect open-vocabulary observations with navigable scene elements~\cite{zhao2024overnav}. 
3D-Mem~\cite{yang2025mem} combines incremental 3D scene memory with informative visual observations, and hierarchical semantic--geometric maps explicitly bridge visual and spatial representations~\cite{li2026hsgm}. 
Other systems retain fine-grained visual hierarchies and update memory through reflection~\cite{ge2026evomemnav}, or compress long visual histories for efficient reuse~\cite{hu2026astramemory}. 
These methods emphasize spatial organization, visual fidelity, and selective retrieval. RetroNav complements them by separating environmental evidence from task-dependent judgment and linking observations, poses, transitions, and views to regions, places, and objects. Historical interpretations and actions remain experience records, so reflection can inspect past decisions without treating them as scene facts.

\textbf{Training-free agent navigation.}
Training-free navigation uses frozen foundation models together with prompts, tools, or pretrained controllers, avoiding additional optimization of navigation-task parameters. TANGO~\cite{ziliotto2025tango} composes language-model reasoning with reusable embodied primitives for zero-shot open-world tasks. Open-Nav~\cite{qiao2025opennav}, AgenticNav~\cite{li2026agenticnav}, and HarnessVLN~\cite{chen2026harnessvln} organize perception, planning, tool use, grounding, and recovery around multimodal language models. These approaches demonstrate strong zero-shot execution, but their primary focus is single-task control or the organization of local history. RetroNav targets their intersection with lifelong navigation: fixed model parameters, persistent scene-memory consolidation, independent post-attempt evaluation, and evidence-grounded retry reflection are combined into one closed loop.

\section{Method}
\label{sec:method}

\subsection{Problem Setting and Overview}
\label{sec:setting}

We consider an agent performing a sequence of navigation tasks $\{\mathcal{T}_i\}_{i=1}^{N}$ in a persistent scene. Each task $\mathcal{T}_i=(g_i,s_i^0)$ specifies a goal and an initial pose. The goal may be a language instruction, an object category or description, or a reference image. At each decision step, the agent receives RGB-D observations and pose information, and selects a movement action or STOP. 
A navigation attempt ends upon STOP or budget exhaustion. A retry addresses the same goal from its original starting pose, while a subsequent task introduces a new goal and its specified start. Memory persists within the scene and task-sequence boundaries of the evaluation protocol.

RetroNav couples online navigation with memory accumulation and scene-grounded reflection (\cref{fig:pipeline}). During navigation, a View Selection Agent chooses a direction and a Waypoint Selection Agent selects a geometrically feasible destination. 
After each attempt, the Memory Builder update the unified memory and an independent evaluator accesses whether the attempt succeeded. If the attempt is judged unsuccessful, a Retrospection Agent reviews historical decisions against the accumulated scene evidence and generates a retry brief. The robot then restarts navigation from its initial pose, guided by the brief.

For a single attempt, we omit task and attempt indices. Let $g$ denote the goal, $s_0$ the initial pose, $\mathcal{M}$ the available unified memory, and $b$ the retry brief. The first attempt has no retry brief. Navigation produces an attempt trace
\begin{equation}
    \tau=\operatorname{Navigate}(g,s_0;\mathcal{M},b).
    \label{eq:attempt}
\end{equation}
Scene knowledge accumulates across tasks, whereas the brief revises the decision context for another attempt at the same task.

\subsection{Unified Memory}
\label{sec:memory}

\begin{figure*}[t]
    \centering
    \includegraphics[width=\textwidth]{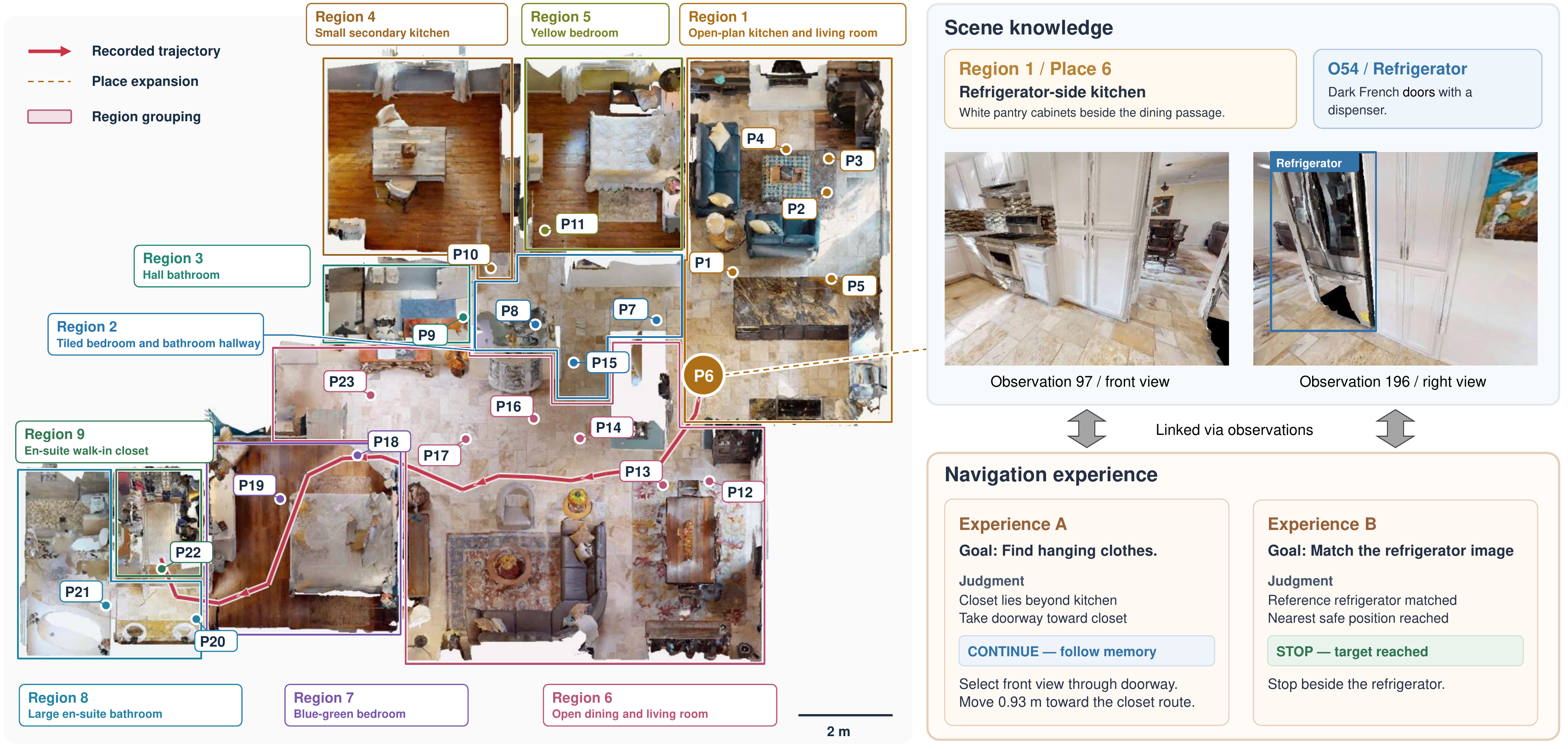}
    \caption{Structure and organization of the unified memory. Left: Cumulative scene knowledge acquired across navigation tasks, including regions, places and trajectories. Top right: Representative scene knowledge at Place 6 of Region 1. Bottom right: Navigation experiences at the same place across distinct tasks, recording different goals, judgments, and actions.}
    \label{fig:scene_memory}
\end{figure*}

A single trajectory yields both physical evidence and task reasoning. RetroNav structures this history into a unified memory:
\begin{equation}
    \mathcal{M}=(\mathcal{S},\mathcal{H},\mathcal{L}),
    \label{eq:memory}
\end{equation}
where $\mathcal{S}$ represents scene knowledge across space and entities, $\mathcal{H}$ preserves chronological navigation experiences, and $\mathcal{L}$ denotes the observation-centric associations bridging the two tracks.
\cref{fig:scene_memory} illustrates how observations link shared scene knowledge to distinct task-specific judgments and actions at the same place.

\textbf{Core representations.}
The anchor of unified memory is the observation $o$, defined by a 6-DoF pose, multi-view RGB-D images, and its attempt identifier. Observations serve as immutable empirical evidence: semantic classifications may be revised over time, but raw observations remain uncorrupted.

Building upon observations, scene knowledge $\mathcal{S}$ captures cumulative physical and semantic structure:
(a)~\emph{Places} group visits to the same localized vantage, aggregating member observations, keeping representative views, and synthesizing grounded visual descriptions.
(b)~\emph{Regions} denote concrete architectural enclosures (e.g., a specific bedroom), organizing confirmed places while decoupling viewing vantage from observed entities.
(c)~\emph{Objects} represent persistent 3D entities with metric bounding boxes, point clouds, categories, and appearance descriptions.
(d)~\emph{Transitions} record verified traversability derived exclusively from executed trajectory endpoints.

Complementary to scene knowledge, navigation experience $\mathcal{H}$ archives subjective decision history. 
The attempt trace is defined as 
\begin{equation}
    \tau=\left(g,\{(o_t,z_t,a_t)\}_{t=0}^{T}\right),
    \label{eq:experience}
\end{equation}
where $o_t$ is the observation at step $t$, $z_t$ records explicit task interpretations and decision rationales, and $a_t$ specifies the chosen action. The experience also retains execution feedback associated with each action. Historical interpretations remain unchanged for subsequent review.

\textbf{Memory construction.}
Upon completing an attempt, the trajectory is processed via two streams:

1.~\emph{Experience organization}: The sequential trace of goals, decisions, rationales, and execution feedback is structured into $\tau$, indexing each decision by its observation $o_t$. Task-specific hypotheses remain quarantined in $\mathcal{H}$, preventing subjective bias from polluting the objective environmental model.

2.~\emph{Hierarchical scene consolidation}: Scene knowledge $\mathcal{S}$ is updated in two stages. First, a vision-language model proposes object bounding boxes and semantic descriptions from RGB images, and SAM~\cite{kirillov2023sam} produces segmentation masks from these boxes (see \cref{sec:supp_memory_updates}). The masks and depth observations are used to back-project object evidence into 3D metric coordinates and fuse it into cumulative 3D object instances via spatial clustering and overlap association~\cite{gu2024conceptgraphs}.
Then, the Memory Builder receives the cumulative Bird's-Eye-View (BEV) map and observations from the current attempt. It clusters observations into places, assigns places to regional instances, and refines object semantic summaries while holding object geometries invariant. Both successful and failed attempts contribute to scene consolidation.

\textbf{Observation-centric dual association.}
The association layer $\mathcal{L}$ connects experiences $\mathcal{H}$ with scene knowledge $\mathcal{S}$ via provenance binding and contextual linking:

1.~\emph{Deterministic provenance chain}: Each decision event is anchored at its observation, which connects it to the corresponding place, supported region membership, and associated object evidence. This establishes the provenance chain \emph{decision $\rightarrow$ observation $\rightarrow$ scene entities}.

2.~\emph{Bidirectional contextual linking}: Rather than storing passive logs, $\mathcal{L}$ enables active cross-track querying. From decision to scene, reviewing a decision at $o_t$ retrieves cumulative multi-perspective evidence accumulated across attempts at that place. From scene to experience, any place or object can be queried to retrieve historical attempts and failures in its vicinity, enabling the agent to contextualize past choices against cumulative spatial facts during reflection.

\subsection{Memory-Assisted Navigation}
\label{sec:navigation}

At each navigation step $t$, we split the decision into two agents. The View Selection Agent dynamically retrieves relevant scene context and cross-task experiences and outputs a moving direction and motion intent. Next, the Waypoint Selection Agent selects an executable waypoint.

\textbf{View Selection.}
At each decision step $t$, the View Selection Agent serves as the semantic planner. It conditions on the navigation goal $g$, four-directional panoramic RGB observations $o_t$, the local metric BEV map $G_t$, the ongoing within-attempt history $h_t$, and the retry brief $b$ (where $b = \emptyset$ for the initial attempt). 

To facilitate the unified memory, the agent dynamically queries $\mathcal{M}$ to construct a task-relevant context $m_t$.
First, using the current robot pose, it indexes historical observation positions in $\mathcal{S}$ to maintain a local context, which contains the region summary, adjacent place descriptions, and visible objects. When this local context is insufficient to select a view, the agent calls \texttt{search\_memory} to retrieve relevant scene entities and uses \texttt{read\_memory} to obtain detailed descriptions or source RGB images for visual verification.

When encountering route branching, landmark ambiguity, or conflicting sensory cues, the agent queries $\mathcal{L}$ using the spatial entities as anchors and accesses relevant decision fragments from other completed tasks. 
Retrieval prioritizes decision events that occurred at that place or explicitly referenced the target entity, filtered by relevance to the ongoing unresolved question. Each retrieved fragment encapsulates the historical goal, task interpretation, observation pose, chosen action, explicit rationale, and subsequent execution consequences.
This observation-centric retrieval provides synthesized memory context $m_t$.

Conditioned on this context, the agent outputs either $\mathrm{STOP}$ or a selected view and motion intent:
\begin{equation}
    \operatorname{View}(g,o_t,G_t,h_t,m_t,b)
    = \begin{cases}
        \mathrm{STOP}, & \text{if stop},\\
        (v_t,c_t), & \text{otherwise}.
    \end{cases}
    \label{eq:view}
\end{equation}
where $v_t$ specifies one of the four cardinal views and $c_t$ denotes the high-level motion intent (e.g., ``traverse through the corridor toward the far doorway'').

To support post-attempt analysis, the agent explicitly records which parts of its historical experiences were used to make each decision. This helps distinguish experiences that actually influenced the decision from those that were only examined. During subsequent reflection (\cref{sec:reflection}), these records allow the agent to check whether past experiences were correctly applied to the current situation or incorrectly generalized.

\textbf{Waypoint selection and execution.}
When the View Selection Agent outputs $(v_t, c_t)$, the Waypoint Selection Agent translates this semantic guidance into a reachable destination. Operating independently of long-term memory $\mathcal{M}$, it focuses strictly on local collision avoidance and physical traversability.

Specifically, we use the depth map of $v_t$ to generate a discrete set of candidate waypoints $\mathcal{W}_t$. The agent evaluates these candidates against the visual context and semantic motion intent to select the optimal waypoint:
\begin{equation}
    w_t = \operatorname{Waypoint}(v_t,\mathcal{W}_t,c_t).
    \label{eq:waypoint}
\end{equation}
After executing $w_t$, the execution feedback and $(o_t, z_t, a_t)$ are appended to the attempt trace $\tau$. 
Then, we acquire the updated observations $o_{t+1}$, local BEV map $G_{t+1}$, and attempt history $h_{t+1}$.
Upon termination, the completed experience enters unified memory construction (\cref{sec:memory}) and post-attempt evaluation (\cref{sec:reflection}).

\subsection{Evaluation and Retrospection}
\label{sec:reflection}

When a navigation attempt ends (either the agent calls $\mathrm{STOP}$ or exhausts its step budget), we use an independent Evaluator Agent to check whether the task is succeeded. 
If the attempt is judged unsuccessful, the system reflects on the failure to identify mistakes and provide guidance for the next retry.

\textbf{Evaluator.}
After an attempt finishes, we first update the scene memory with $\tau$. During navigation, the robot may stop too early or go to the wrong way due to misjudgments. 
Asking the navigator to judge itself usually fails to catch these mistakes. Therefore, we use an independent Evaluator to decide whether the attempt actually succeeded.
The Evaluator takes the goal $g$, the full attempt trace $\tau$, and the updated scene memory $\mathcal{S}$ as input. It evaluates the attempt from two aspects.
First, it checks whether the executed path followed the key landmarks and steps described in the instruction.
Second, it compares the final stopping position with the updated scene memory to confirm whether the robot truly stopped near the correct target, rather than near a confusing or similar object.
Finally, the evaluator outputs a final judgment of success or failure. A retry is triggered only when the attempt is judged as a failure, which prevents blind and unnecessary retries.

\textbf{Retrospection Agent.}
Once a retry is triggered, the Retrospection Agent reviews the failed trajectory $\tau$ to find the reasons behind the failure. In many cases, navigation fails not because the environment lacks information, but because the agent misinterpreted its visual observations.
Based on the accumulated scene memory $\mathcal{S}$, the agent examines three common types of decision errors:
(a)~Goal or landmark misidentification: checking if the agent mistook a similar room or an incorrect object for the target;
(b)~Route selection mistakes: checking if the agent turned in the wrong direction at corridor intersections or doorways, leading into dead ends;
(c)~Experience misapplication: checking the cross-task experiences cited in the decision rationales $z_t$ through the association layer $\mathcal{L}$, to verify if past experiences were incorrectly applied.

During this review, the agent can use the full scene memory $\mathcal{M}$ to clarify confusing visual details. Finally, the Retrospection Agent summarizes these diagnostic results into a structured retry brief $b$:
\begin{equation}
    b = \operatorname{Reflect}(g, \tau, \mathcal{M}).
    \label{eq:reflection}
\end{equation}
The brief contains four key parts:
(a)~Goal clarification: clarifying the true visual features of the goal to correct earlier misunderstandings;
(b)~Key decision reviews: identifying up to three pivotal mistakes in the trajectory, explaining what went wrong and how to correct the spatial intent;
(c)~Route suggestion: recommending a better route while warning against the wrong paths taken before;
(d)~Reference images: selecting up to eight key images from scene memory (such as different views of the target entrance) as visual references for the next attempt.

\textbf{Retry.}
With the retry brief $b$, the robot resets to its initial pose $s_0$ and starts a new attempt.
In the new attempt, the View Selection Agent takes both the brief $b$ and the reference images as part of its input (\cref{sec:navigation}). The brief acts like a checklist: when the robot approaches a location where it previously failed, the brief reminds it not to repeat the past mistake. Meanwhile, the robot still relies on real-time camera views to make step-by-step movement and collision-avoidance decisions. In this way, accumulated scene memory not only benefits future tasks, but also directly helps fix failures in the current task.

\section{Experiments}
\label{sec:experiments}

\begin{table}[t]
\centering
\footnotesize
\setlength{\tabcolsep}{-0.5pt}
\renewcommand{\arraystretch}{0.96}
\caption{Evaluation results on R2R-CE and RxR-CE val-unseen. -- denotes unreported. $\dagger$ indicates a method retested with an updated model. The best results are shown in \textbf{bold}, and the second best is \underline{underlined}.}
\label{tab:r2r_rxr}
\begin{tabular*}{\columnwidth}{@{\extracolsep{\fill}}lcccccc@{}}
\toprule
& \multicolumn{3}{c}{R2R-CE} & \multicolumn{3}{c}{RxR-CE} \\
\cmidrule(lr){2-4}\cmidrule(lr){5-7}
Method & SR & SPL & OSR & SR & SPL & OSR \\
\midrule
\multicolumn{7}{l}{\textit{Training-based}} \\
NavMorph (HNR)~\cite{yao2025navmorph} & \underline{64.0} & \underline{53.0} & \underline{68.0} & 58.0 & 49.0 & -- \\
Uni-NaVid~\cite{zhang2025uninavid} & 47.0 & 42.7 & 53.3 & 48.7 & 40.9 & \textbf{55.5} \\
StreamVLN-Prune-Mem~\cite{wei2026streamvln} & 57.4 & 51.1 & 65.5 & 53.9 & 45.1 & -- \\
DualVLN~\cite{wei2026dualvln} & \textbf{64.3} & \textbf{58.5} & \textbf{70.7} & \underline{61.4} & \underline{51.8} & -- \\
ReflectVLN~\cite{wang2026reflectvln} & 62.8 & \textbf{58.5} & 67.3 & \textbf{66.0} & \textbf{57.2} & -- \\
\midrule
\multicolumn{7}{l}{\textit{Training-free}} \\
GC-VLN (DeepSeek-R1)~\cite{yin2025gcvln} & 33.6 & 16.3 & 41.8 & 33.8 & 13.8 & 44.4 \\
HSGM (GPT-5)~\cite{li2026hsgm} & 47.9 & 32.8 & 58.7 & 41.8 & 25.1 & -- \\
HarnessVLN (GPT-5.5)~\cite{chen2026harnessvln} & 59.6 & 43.6 & 70.7 & 51.4 & 38.1 & -- \\
Open-Nav (Gemini-2.5-Pro)~\cite{qiao2025opennav} & 27.0 & 23.2 & 32.0 & -- & -- & -- \\
AgenticNav (Gemini-3.7-Flash)~\cite{li2026agenticnav} & 76.0 & \underline{66.5} & 78.0 & -- & -- & -- \\
HAM-VLN (Gemini-3.1-Pro)~\cite{liu2026hamvln} & 61.0 & 48.1 & 78.7 & 52.7 & 35.7 & -- \\
SparseNav (GPT-5)~\cite{chen2026sparsenav} & 42.8 & 35.2 & 53.4 & 40.7 & 24.1 & -- \\
Uni-LaViRA (Gemini-3.1-Pro)~\cite{ding2026unilavira} & 60.7 & 47.7 & 73.7 & 51.3 & 34.0 & -- \\
Uni-LaViRA (GPT-5.6-Sol)$^{\dagger}$~\cite{ding2026unilavira} & 47.2 & 32.2 & 57.3 & 52.4 & 34.4 & 69.1 \\
Ours (GPT-6-Astra) & \textbf{84.0} & \textbf{75.4} & \textbf{87.7} & \textbf{84.0} & \textbf{70.9} & \textbf{90.3} \\
Ours (GPT-5.6-Sol) & \underline{76.3} & 66.4 & \underline{80.7} & \underline{76.7} & \underline{61.0} & \underline{80.0} \\
Ours (Qwen3.8-27B) & 58.0 & 49.5 & 67.3 & 64.0 & 46.4 & 71.0 \\
\bottomrule
\end{tabular*}
\end{table}

\begin{table}[t]
\centering
\footnotesize
\setlength{\tabcolsep}{1.0pt}
\renewcommand{\arraystretch}{0.96}
\caption{Evaluation results on HM3D-OVON val-unseen. -- denotes unreported. The best results are shown in \textbf{bold}, and the second best is \underline{underlined}.}
\label{tab:ovon}
\begin{tabular*}{\columnwidth}{@{\extracolsep{\fill}}lccc@{}}
\toprule
Method & SR & SPL & OSR \\
\midrule
\multicolumn{4}{l}{\textit{Training-based}} \\
Uni-NaVid~\cite{zhang2025uninavid} & 39.5 & \underline{19.8} & -- \\
AstraNav-Memory~\cite{hu2026astramemory} & \textbf{62.5} & \textbf{34.9} & -- \\
MTU3D~\cite{zhu2025mtu3d} & \underline{40.8} & 12.1 & -- \\
\midrule
\multicolumn{4}{l}{\textit{Training-free}} \\
SpaceVLN (Qwen3.5-Plus/Flash)~\cite{deng2026spacevln} & 51.6 & 33.1 & -- \\
HarnessVLN (GPT-5.6-Luna)~\cite{chen2026harnessvln} & 59.3 & 36.6 & -- \\
TANGO (GPT-4o)~\cite{ziliotto2025tango} & 35.5$\pm$0.3 & 19.5$\pm$0.3 & -- \\
MetaNav (GPT-4o)~\cite{li2026metanav} & 46.1 & 29.8 & -- \\
DRIVE-Nav (Qwen3-VL-8B)~\cite{gao2026drivenav} & 50.2 & 32.6 & -- \\
AECNav (C-RADIOv4)~\cite{liu2026aecnav} & 57.3 & 30.5 & -- \\
Ours (GPT-6-Astra) & \textbf{66.9$\pm$1.2} & \textbf{50.2$\pm$1.4} & \textbf{71.3$\pm$1.2} \\
Ours (GPT-5.6-Sol) & \underline{60.1$\pm$1.8} & 44.5$\pm$2.1 & \underline{65.1$\pm$1.8} \\
Ours (Qwen3.8-27B) & 55.4$\pm$3.3 & \underline{45.8$\pm$2.5} & 59.8$\pm$3.3 \\
\bottomrule
\end{tabular*}
\end{table}

\begin{table}[t]
\centering
\footnotesize
\setlength{\tabcolsep}{0pt}
\renewcommand{\arraystretch}{0.96}
\caption{Evaluation results on IR2R-CE val-unseen. $\dagger$ indicates a method retested with an updated model. The best results are shown in \textbf{bold}, and the second best is \underline{underlined}.}
\label{tab:ir2r}
\begin{tabular*}{\columnwidth}{@{\extracolsep{\fill}}lccc@{}}
\hline
Method & SR & SPL & t-nDTW \\
\hline
\multicolumn{4}{@{}l}{\textit{(a) Training-based}} \\
CMA~\cite{krantz2020vlnce} & 19.0$\pm$1.0 & 18.0$\pm$1.0 & 38.0$\pm$2.0 \\
TourCMA~\cite{krantz2023ivln} & 18.0$\pm$0.0 & 17.0$\pm$1.0 & 36.0$\pm$1.0 \\
PoolCMA~\cite{krantz2023ivln} & 16.0$\pm$1.0 & 15.0$\pm$0.0 & 36.0$\pm$2.0 \\
PoolEndCMA~\cite{krantz2023ivln} & 18.0$\pm$1.0 & 16.0$\pm$1.0 & 38.0$\pm$2.0 \\
OVER-NAV (iterative)~\cite{zhao2024overnav} & \underline{35.0$\pm$1.0} & \underline{33.0$\pm$1.0} & \underline{50.0$\pm$2.0} \\
ETPNav~\cite{an2024etpnav} & 28.0$\pm$1.0 & 27.0$\pm$1.0 & 41.0$\pm$2.0 \\
HNR~\cite{wang2024hnr} & 30.0$\pm$1.0 & 28.0$\pm$2.0 & 44.0$\pm$2.0 \\
SeqWalker~\cite{han2026seqwalker} & \textbf{36.0$\pm$2.0} & \textbf{34.0$\pm$2.0} & \textbf{52.0$\pm$2.0} \\
\hline
\multicolumn{4}{@{}l}{\textit{(b) Training-free}} \\
CA-Nav (GPT-5.6-Sol)$^{\dagger}$~\cite{chen2025canav} & 52.2$\pm$0.5 & 44.7$\pm$0.9 & 47.6$\pm$0.9 \\
Open-Nav (GPT-5.6-Sol)$^{\dagger}$~\cite{qiao2025opennav} & 62.2$\pm$0.6 & 51.4$\pm$1.0 & 49.2$\pm$0.2 \\
NavHarness (Qwen3.8-27B)~\cite{zhao2026navharness} & 45.1$\pm$1.1 & 32.0$\pm$0.9 & 43.2$\pm$1.2 \\
NavHarness (Opus 5)~\cite{zhao2026navharness} & 78.2$\pm$0.9 & 56.9$\pm$1.3 & 58.1$\pm$1.0 \\
NavHarness (GPT-6-Astra)~\cite{zhao2026navharness} & \underline{85.9$\pm$0.8} & \textbf{76.1$\pm$0.9} & 64.2$\pm$1.1 \\
Ours (GPT-6-Astra) & \textbf{89.2$\pm$0.2} & \underline{74.5$\pm$0.6} & \textbf{78.8$\pm$0.1} \\
Ours (GPT-5.6-Sol) & 84.1$\pm$0.2 & 71.9$\pm$1.0 & \underline{75.2$\pm$0.0} \\
Ours (Qwen3.8-27B) & 57.2$\pm$7.1 & 48.1$\pm$4.8 & 68.4$\pm$1.1 \\
\hline
\end{tabular*}
\end{table}

\begin{table}[t]
\centering
\footnotesize
\setlength{\tabcolsep}{1.5pt}
\caption{Evaluation results on GOAT-Bench val-unseen. -- denotes unreported. The best results are shown in \textbf{bold}, and the second best is \underline{underlined}.
}
\label{tab:goat}
\begin{tabular}{@{}lccc@{}}
\toprule
Method & SR & SPL & e-SR \\
\midrule
\multicolumn{4}{@{}l}{\textit{(a) Training-based}}\\
SenseAct-NN Skill Chain~\cite{khanna2024goat} & 29.5 & 11.3 & -- \\
SenseAct-NN Monolithic~\cite{khanna2024goat} & 12.3 & 6.8 & -- \\
AstraNav-Memory~\cite{hu2026astramemory} & \textbf{62.7} & \textbf{56.9} & -- \\
MTU3D~\cite{zhu2025mtu3d} & \underline{47.2} & 27.7 & -- \\
MemoryExplorer (LMEE)~\cite{wang2026lmee} & 46.4 & \underline{28.0} & -- \\
\midrule
\multicolumn{4}{@{}l}{\textit{(b) Training-free}}\\
TANGO (GPT-4o)~\cite{ziliotto2025tango} & 32.1 & 16.5 & -- \\
SSMG-Nav (Qwen-VL-Plus)~\cite{niu2026ssmg} & 46.5 & 34.1 & 8.6 \\
EvoMemNav (Qwen3-VL-8B)~\cite{ge2026evomemnav} & 59.6 & 38.9 & -- \\
STEGNav (GPT-5.4-mini)~\cite{chen2026stegnav} & 66.3 & 39.7 & -- \\
MSGNav (GPT-4o, 0.25\,m)~\cite{huang2026msgnav} & 52.0 & 29.6 & -- \\
3D-Mem (GPT-4o)~\cite{yang2025mem} & 69.1 & 48.9 & -- \\
HIMM (GPT-4o)~\cite{li2026himm} & 72.8 & 56.1 & -- \\
MetaNav (GPT-4o)~\cite{li2026metanav} & 71.4 & 51.8 & -- \\
NavHarness (Qwen3.8-27B)~\cite{zhao2026navharness} & 71.7$\pm$0.5 & 48.2$\pm$0.4 & 14.4$\pm$0.7 \\
NavHarness (Opus5)~\cite{zhao2026navharness} & 81.5$\pm$0.7 & 55.0$\pm$0.6 & 28.6$\pm$0.8 \\
NavHarness (GPT-6-Astra)~\cite{zhao2026navharness} & \underline{83.7$\pm$0.4} & \textbf{62.3$\pm$0.3} & \underline{36.9$\pm$0.9} \\
Ours (GPT-6-Astra) & \textbf{84.1$\pm$1.2} & \underline{58.4$\pm$0.8} & \textbf{40.3$\pm$3.1} \\
Ours (GPT-5.6-Sol) & 82.3$\pm$0.5 & 57.6$\pm$1.5 & 35.3$\pm$1.3 \\
Ours (Qwen3.8-27B) & 74.4$\pm$5.3 & 45.7$\pm$3.6 & 17.1$\pm$2.9 \\
\bottomrule
\end{tabular}
\end{table}

\subsection{Experimental Setup}
\label{sec:exp_setup}

\textbf{Benchmarks and task settings.}
We evaluate our framework across five standard embodied navigation benchmarks: R2R-CE~\cite{anderson2018vln,krantz2020vlnce}, RxR-CE~\cite{ku2020rxr,krantz2020vlnce}, HM3D-OVON~\cite{yokoyama2024ovon}, IR2R-CE~\cite{krantz2023ivln}, and GOAT-Bench~\cite{khanna2024goat}.
Because R2R-CE, RxR-CE, and HM3D-OVON are episodic benchmarks that reset between tasks, we disable cross-task Unified Memory and evaluate standard navigation with failure reflection and retry. In contrast, for benchmarks with persistent environments (IR2R-CE and GOAT-Bench), Unified Memory is enabled to accumulate scene knowledge and experience across consecutive episodes.

\textbf{Foundation models.}
To evaluate performance across different model families and scales, we deploy our training-free framework with three foundation models: GPT-6-Astra~\cite{openai2026astra}, GPT-5.6-Sol~\cite{openai2026sol}, and the open-source Qwen3.8-27B~\cite{qwen2026modelcard}.
For all experiments, 2D object candidate extraction is performed by GPT-5.6-Terra, instance segmentation is produced by SAM~\cite{kirillov2023sam}.

\textbf{Evaluation scale and sampling protocol.}
Due to the substantial computational cost and API invocation overhead of frontier multimodal agents, most training-free navigation methods evaluate on representative benchmark subsets~\cite{qiao2025opennav,dai2026evonav,li2026agenticnav,ding2026unilavira,liu2026hamvln}. 
Following this established protocol, we evaluate on: (a) 100 episodes for R2R-CE, aligned with Open-Nav~\cite{qiao2025opennav}; (b) 100 randomly sampled episodes from the RxR-CE val-unseen split with English (en-US) guide instructions; (c) 150 subtasks for HM3D-OVON; (d) 299 episodes across 11 tours for IR2R-CE; and (e) 307 multimodal subtasks from val-unseen for GOAT-Bench. 
As demonstrated in Uni-LaViRA~\cite{ding2026unilavira}, subset evaluations yield performance trends highly consistent with full benchmarks.
Main results are means of three independent runs.

\textbf{Evaluation metrics.}
Across all benchmarks, we report standard embodied navigation metrics: Success Rate (SR) measuring goal reach criteria, and Success weighted by Path Length (SPL) penalizing trajectory inefficiency. For continuous instruction following (R2R-CE, RxR-CE, HM3D-OVON), we also report Oracle Success Rate (OSR), indicating whether the agent visited the goal threshold at any step. For multi-tour IR2R-CE, we report tour-level normalized Dynamic Time Warping (t-nDTW) to assess fidelity to the demonstration paths across successive episodes. For GOAT-Bench, we report subtask-level SR and SPL, as well as Episode Success Rate (e-SR), which strictly requires every subtask in an episode sequence to succeed.
SPL in the main tables is computed from each task's final attempt. We additionally evaluate path efficiency using the total distance traveled across all attempts in \cref{sec:supp_cumulative_spl}.

\subsection{Episodic Navigation}
\label{sec:exp_episodic}

We evaluate navigation performance under standard episodic configurations (R2R-CE, RxR-CE, and HM3D-OVON) where environments reset between tasks and cross-task memory is disabled. In this setting, the agent relies solely on online perception, planning, and our failure reflection and retry mechanism. 

As shown in \cref{tab:r2r_rxr,tab:ovon}, RetroNav consistently outperforms existing training-free methods across both instruction-following and open-vocabulary object-goal search. 
Notably, equipped with GPT-6-Astra, RetroNav establishes new state-of-the-art results on R2R-CE (84.0\% SR) and RxR-CE (84.0\% SR), substantially surpassing the training-free AgenticNav (76.0\% SR on R2R-CE) and the training-based ReflectVLN (66.0\% SR on RxR-CE).
Similarly, in open-vocabulary object search on HM3D-OVON, RetroNav achieves a leading 66.9\% SR and 50.2\% SPL, outperforming existing approaches.
Crucially, this strong performance generalizes across model families: equipped with GPT-5.6-Sol, RetroNav achieves 76.3\% SR on R2R-CE, ranking second and outperforming AgenticNav (76.0\% SR), while open-source Qwen3.8-27B also achieves solid success rates and efficiency on R2R-CE and RxR-CE, demonstrating that our reflection and retry mechanism effectively recovers from execution errors regardless of the underlying foundation model.

\subsection{Navigation in Persistent Environments}
\label{sec:exp_persistent}

\begin{figure*}[t]
\centering
\includegraphics[width=\textwidth]{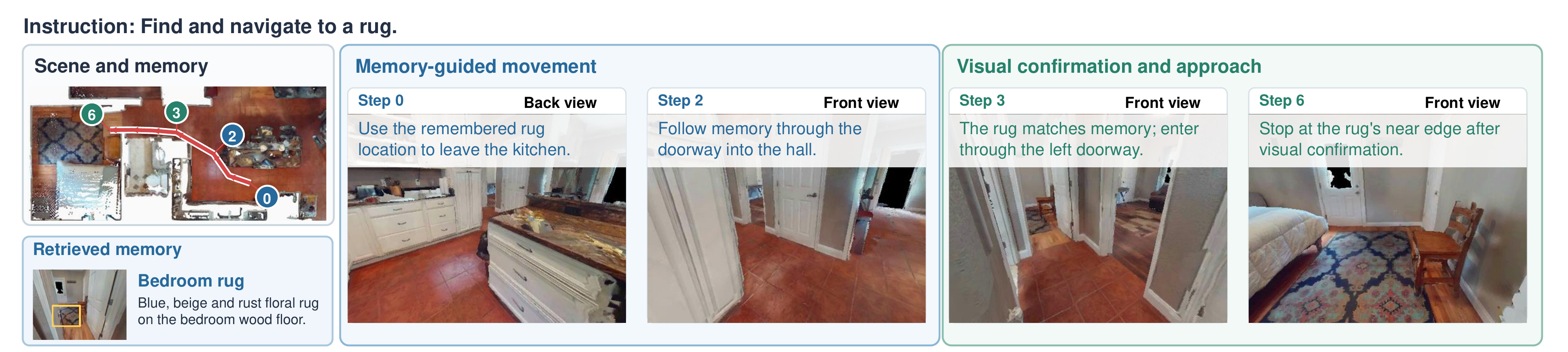}
\caption{Qualitative trace of memory-guided navigation. Given ``\textit{Find and navigate to a rug}'', RetroNav retrieves room context and visual cues from Unified Memory $\mathcal{M}$, steering the agent out of the kitchen (Steps~0--2) to locate and stop at the rug (Steps~3--6).}
\label{fig:memory_nav}
\end{figure*}

We next evaluate lifelong navigation in persistent environments (IR2R-CE and GOAT-Bench), where our full Unified Memory module is active to accumulate spatial knowledge and past navigation experiences across successive episodes. 

As summarized in \cref{tab:ir2r,tab:goat}, retaining and grounding decisions on cumulative scene memory yields marked performance gains. On multi-tour IR2R-CE, RetroNav with GPT-6-Astra achieves a leading 89.2\% SR and an unprecedented 78.8\% t-nDTW, demonstrating that cumulative spatial awareness not only helps the agent reach destination goals but also substantially improves adherence to human demonstration paths across repeated visits. 

Furthermore, on GOAT-Bench's demanding multimodal sequential navigation, RetroNav with GPT-6-Astra maintains a high subtask SR of 84.1\% while achieving a remarkable 40.3\% episode success rate (e-SR), surpassing NavHarness (36.9\% e-SR).
 Notably, open-source Qwen3.8-27B achieves a solid 74.4\% subtask SR and 17.1\% e-SR, outperforming established open-source baselines such as EvoMemNav (59.6\% SR) and SSMG-Nav (8.6\% e-SR). 
 These results confirm that our Unified Memory architecture effectively transforms historical observations and reflection outcomes into reusable environmental assets, significantly reducing compounding errors in long-horizon embodied tasks across diverse model families. 
 As illustrated in \cref{fig:memory_nav}, when instructed to locate a target object (e.g., a rug), RetroNav retrieves the target's spatial anchor and visual cues from Unified Memory $\mathcal{M}$, navigating directly toward the destination without blind exploration.

\subsection{Ablation Study}
\label{sec:exp_ablation}

To understand the individual contributions of our core design choices, we conduct systematic ablations across persistent and episodic benchmarks using GPT-5.6-Sol.

\begin{table}[t]
\centering
\footnotesize
\setlength{\tabcolsep}{0pt}
\caption{Component-wise ablation study on IR2R-CE and GOAT-Bench. We evaluate the contributions of Naive Retry (Naive), Unified Memory ($\mathcal{M}$), Evaluator (Eval.), and Retry Brief (Brief).}
\label{tab:ablation_components}
\begin{tabular*}{\columnwidth}{@{\extracolsep{\fill}}ccccccccccc@{}}
\toprule
& \multicolumn{4}{c}{Component} & \multicolumn{3}{c}{IR2R-CE} & \multicolumn{3}{c}{GOAT-Bench} \\
\cmidrule(lr){2-5}\cmidrule(lr){6-8}\cmidrule(lr){9-11}
Row & \makebox[2.0em]{Naive} & \makebox[1.6em]{$\mathcal{M}$} & \makebox[1.8em]{Eval.} & \makebox[2.0em]{Brief} & \makebox[2.2em]{SR} & \makebox[2.2em]{SPL} & \makebox[3.0em]{t-nDTW} & \makebox[2.2em]{SR} & \makebox[2.2em]{SPL} & \makebox[2.4em]{e-SR} \\
\midrule
(1) & -- & -- & -- & -- & 65.6 & 57.0 & 68.6 & 67.0 & 41.8 & 14.3 \\
(2) & \checkmark & -- & -- & -- & 76.7 & 62.6 & 71.6 & 81.0 & 47.8 & 28.6 \\
(3) & -- & \checkmark & -- & -- & 69.4 & 59.7 & 74.2 & 71.5 & 49.6 & 21.4 \\
(4) & -- & \checkmark & \checkmark & -- & 74.5 & 63.8 & 75.0 & 75.5 & 51.8 & 26.0 \\
(5) & -- & \checkmark & \checkmark & \checkmark & \textbf{84.1} & \textbf{71.9} & \textbf{75.2} & \textbf{82.3} & \textbf{57.6} & \textbf{35.3} \\
\bottomrule
\end{tabular*}
\end{table}

\begin{figure*}[!t]
\centering
\includegraphics[width=\textwidth]{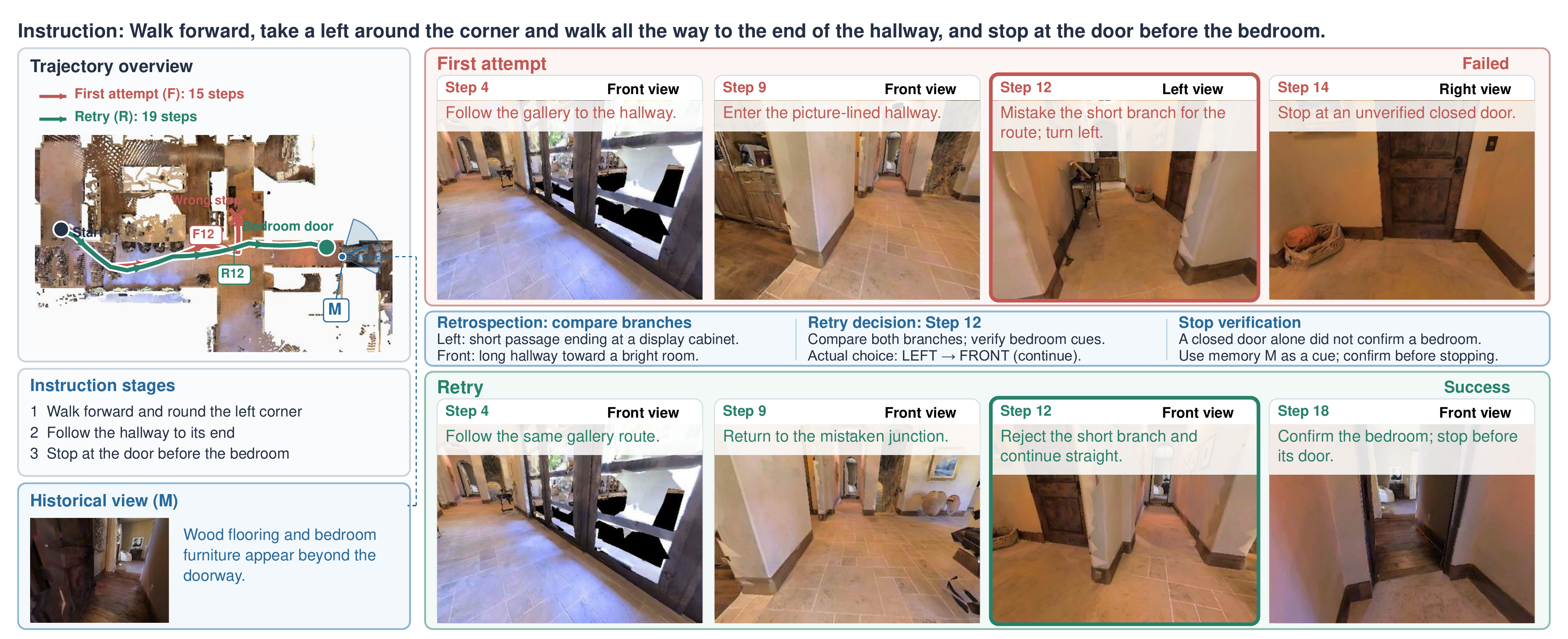}
\caption{Failure retrospection and retry on R2R-CE. After the initial attempt branches into a dead end at Step~12 and stops prematurely (top), the Retry Brief isolates the branch confusion (middle), guiding the retry to bypass the error and reach the bedroom door (bottom).}
\label{fig:reflection}
\end{figure*}

\textbf{Component-level ablation.}
\cref{tab:ablation_components} reports navigation performance across five component configurations on IR2R-CE and GOAT-Bench.
First, comparing Row~(1) and Row~(3) isolates the foundational role of persistent Unified Memory $\mathcal{M}$ under single-attempt execution. Reusing spatial knowledge across tours raises SR from 65.6\% to 69.4\% and lifts trajectory fidelity (t-nDTW from 68.6\% to 74.2\%) on IR2R-CE, while increasing continuous episode completion (e-SR) from 14.3\% to 21.4\% on GOAT-Bench.
Second, comparing unguided retry with causal reflection reveals the critical value of structured remediation. While Naive Retry in Row~(2) recovers a subset of simple execution slips by repetitively exploring (76.7\% SR on IR2R-CE and 81.0\% SR on GOAT-Bench), it achieves only 28.6\% e-SR on GOAT-Bench. Similarly, re-executing with memory but without a Retry Brief in Row~(4) caps episode completion at 26.0\% e-SR, as the agent frequently repeats identical junction mistakes without causal guidance.
Finally, integrating Unified Memory, the Evaluator, and the Retry Brief in Row~(5) achieves the strongest performance across all metrics (84.1\% SR on IR2R-CE and 35.3\% e-SR on GOAT-Bench). This demonstrates that the gains of RetroNav stem from grounded causal retrospection rather than simply granting extra execution budget. As illustrated in \cref{fig:reflection}, when the initial attempt diverges into a deceptive side corridor and stops prematurely, the Retry Brief diagnoses the branching mistake, guiding the retry to reuse the valid prefix, reject the misleading turn, and reach the target door.

\begin{figure*}[!t]
\centering
\includegraphics[width=\textwidth]{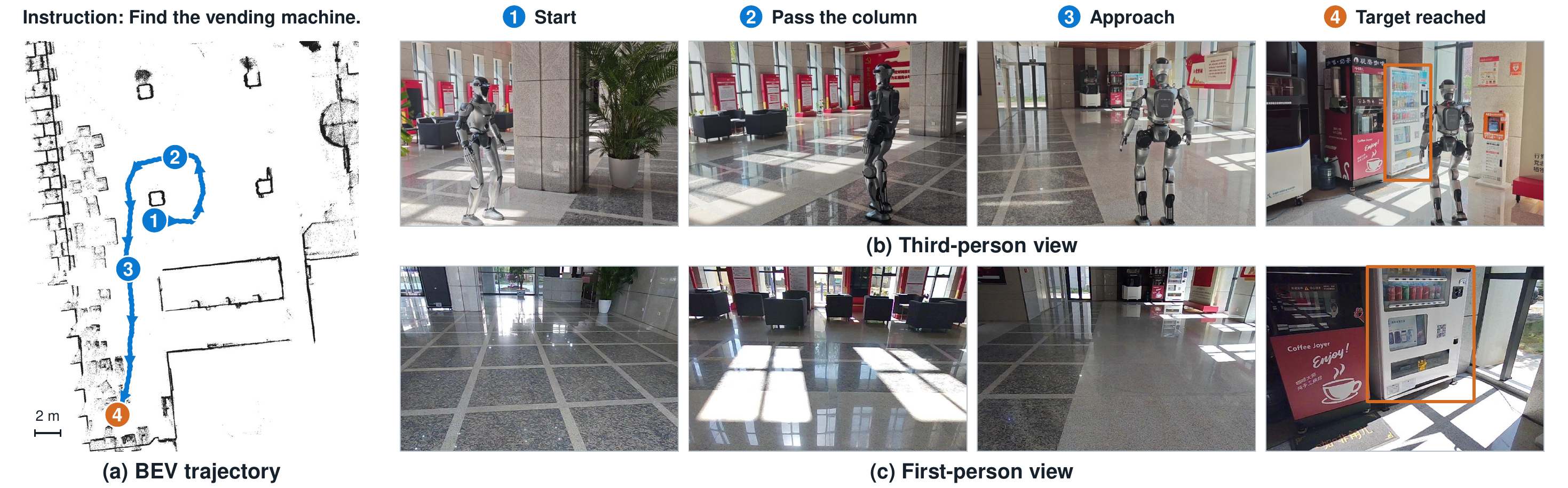}
\caption{Real-world navigation with RetroNav on a CASBOT~02 humanoid robot. Given the instruction ``Find the vending machine,'' the robot navigates through a lobby and reaches the target autonomously. (a) Bird's-eye view of the recorded trajectory. (b,c) Third-person and onboard views at four numbered stages. Orange boxes highlight the target vending machine.}
\label{fig:real_robot}
\end{figure*}

\begin{table}[t]
\centering
\small
\setlength{\tabcolsep}{0pt}
\caption{Ablation of the Retry Brief on 33 fixed first-attempt failures from R2R-CE, without scene memory. Each variant receives one retry from the same first-attempt traces. Rec. denotes retry recovery rate (\%). Checkmarks indicate enabled components.}
\label{tab:ablation_retry_brief}
\begin{tabular*}{\columnwidth}{@{\extracolsep{\fill}}ccccc@{}}
\toprule
Row & Route & Review & Images & Rec. \\
\midrule
(1) & -- & -- & -- & 12.1 \\
(2) & \checkmark & \checkmark & -- & 21.2 \\
(3) & \checkmark & -- & \checkmark & 18.2 \\
(4) & \checkmark & \checkmark & \checkmark & \textbf{25.8} \\
\bottomrule
\end{tabular*}
\end{table}

\textbf{Design of the Retry Brief.}
With unified memory disabled, we evaluate Retry Brief variants on 33 R2R-CE trajectories whose first attempts all failed. Each variant is evaluated twice on the same 33 tasks.
\cref{tab:ablation_retry_brief} summarizes the results.
Blind retry in Row~(1) achieves a 12.1\% recovery rate.
Row~(2) outperforms Row~(3), suggesting that key-decision reviews provide stronger corrective guidance than reference images without such reviews.
By explaining why a previous choice was wrong, decision reviews help the navigator revise mistaken interpretations that route suggestions and images alone may leave unchanged.
The complete Retry Brief in Row~(4) achieves the highest recovery rate of 25.8\%, exceeding blind retry by 13.6 percentage points. 
This improvement can be attributed to the complementary roles of decision reviews and reference images: the reviews identify which decisions need correction, while the images provide visual cues that help the navigator carry out those corrections during the retry.

\begin{table}[t]
\centering
\small
\setlength{\tabcolsep}{0pt}
\caption{Failure-detection quality on R2R-CE. We compare navigator self-stopping, our Evaluator, and the Oracle upper bound.}
\label{tab:ablation_evaluator}
\begin{tabular*}{\columnwidth}{@{\extracolsep{\fill}}lccc@{}}
\toprule
Method & Succ. Recall & Fail. Recall & Accuracy \\
\midrule
Self-Stopping & 100.0 & 0.0 & 55.6 \\
Evaluator (Ours) & 76.0 & 65.0 & 71.1 \\
Oracle & 100.0 & 100.0 & 100.0 \\
\bottomrule
\end{tabular*}
\end{table}

\textbf{Effectiveness of the Evaluator.}
\cref{tab:ablation_evaluator} evaluates failure detection quality separately from downstream retry on R2R-CE. Relying on navigator self-stopping fails to identify unsuccessful episodes due to widespread premature stopping, yielding a 0.0\% failure recall and only 55.6\% accuracy on held-out trajectories. In contrast, our Evaluator achieves 76.0\% recall on successful trajectories, 65.0\% recall on failed trajectories, and 71.1\% overall accuracy without exposing privileged simulator state. This demonstrates that reliable failure detection can trigger targeted retries autonomously in realistic deployments.

\subsection{Real-World Deployment}
\label{sec:exp_real_world}

We deploy RetroNav on a CASBOT~02 humanoid robot\footnote{\url{https://casbot.tech/product/02}}, using a chest-mounted RGB-D camera to obtain egocentric observations and FAST-LIO2~\cite{xu2022fastlio2} to estimate robot poses. At each navigation step, the navigation module selects a waypoint in the visual observation. Using depth, camera intrinsics and extrinsics, and the estimated robot pose, the waypoint is transformed into the robot coordinate frame and passed to the robot's existing locomotion controller for execution. New observations after execution inform subsequent navigation decisions, forming a closed loop of perception, decision-making, and motion execution.

\cref{fig:real_robot} shows RetroNav completing an autonomous object-goal navigation task in a real lobby. The task starts with empty scene memory. Given the instruction ``Find the vending machine,'' the robot moves past a column, approaches the target vending machine, and autonomously determines when to stop upon arrival, without human intervention throughout the task. The bird's-eye trajectory and corresponding third-person and onboard views show the complete navigation process of RetroNav on a real robot.

\section{Conclusion}
\label{sec:conclusion}

In this paper, we presented \textbf{RetroNav}, a training-free self-evolving navigation framework that integrates navigation, memory accumulation, and failure reflection.
RetroNav maintains a unified memory that links raw environmental observations with historical navigation experiences, allowing the agent to re-examine prior choices while keeping factual observations distinct from subjective interpretations.
After each navigation attempt, an independent evaluator assesses task completion. If the attempt is judged unsuccessful, a reflection module reviews prior decisions against accumulated scene evidence and generates a retry brief to guide the next attempt. 
By bridging cross-task knowledge reuse with within-task decision revision, RetroNav achieves state-of-the-art performance across five navigation benchmarks.

Despite its effectiveness, several limitations remain. First, the accuracy of memory construction and failure reflection depends on the quality of upstream perception, spatial association, and model judgment.
Perceptual ambiguities can occasionally misguide the diagnosis. 
Second, while the retry mechanism substantially improves overall task success, additional attempts inevitably incur extra travel steps and inference costs. Future work could investigate adaptive retry policies that balance recovery benefits against execution budgets, as well as more robust cross-task association under sensor noise.

{\small
\bibliographystyle{ieeenat_fullname}
\bibliography{main}
}
\clearpage
\setcounter{page}{1}
\maketitlesupplementary

\section{Evaluation Protocols}
\label{sec:supp_eval_protocols}

\begin{table}[t]
    \centering
    \footnotesize
    \setlength{\tabcolsep}{0pt}
    \caption{Evaluation subsets, cross-task memory, and navigation step budgets across the five benchmarks.}
    \label{tab:supp_eval_protocols}
    \begin{tabular*}{\columnwidth}{@{\extracolsep{\fill}}lccc@{}}
        \toprule
        Benchmark & \shortstack{Evaluation\\size} & \shortstack{Cross-task\\memory} & \shortstack{Max. steps\\per attempt} \\
        \midrule
        R2R-CE & 100 episodes & Disabled & 30 \\
        RxR-CE & 100 episodes & Disabled & 50 \\
        HM3D-OVON & 150 episodes & Disabled & 40 \\
        IR2R-CE & 299 episodes & Enabled & 30 \\
        GOAT-Bench & 307 subtasks & Enabled & 40 \\
        \bottomrule
    \end{tabular*}
\end{table}

\textbf{Evaluation subsets.}
\cref{tab:supp_eval_protocols} summarizes the evaluation settings. All subsets are drawn from the corresponding val-unseen splits. For R2R-CE, we use the 100-episode subset adopted by Open-Nav~\cite{qiao2025opennav}. The IR2R-CE subset contains 299 episodes organized into 11 tours across three scenes. The GOAT-Bench evaluation set contains 307 subtasks across five scenes.

\textbf{Memory accumulation.}
R2R-CE, RxR-CE, and HM3D-OVON are evaluated without cross-task memory accumulation: each new task starts without memory from previous tasks. For IR2R-CE, memory accumulates across successive episodes and tours within the same scene. For GOAT-Bench, memory accumulates across subtasks and composite episodes within the same scene. In both persistent settings, completed attempts update the scene's memory before subsequent tasks use it. Each scene maintains a separate memory, and information is not shared across scenes.

\textbf{Navigation budgets.}
The step limits in \cref{tab:supp_eval_protocols} count navigation decision steps within each attempt. Each attempt ends when the navigator issues STOP or reaches its step limit. A retry starts from the task's original starting pose and receives the same step budget as the first attempt.

\section{Memory Storage, Updates, and Queries}
\label{sec:supp_memory}

While \cref{sec:memory,sec:navigation,sec:reflection} formulate the conceptual principles of the unified memory $\mathcal{M} = (\mathcal{S}, \mathcal{H}, \mathcal{L})$, this section provides the underlying implementation specifications, operational budgets, and empirical evolution details. We structure this analysis along the complete memory lifecycle:
how multi-source records are anchored and represented (\cref{sec:supp_memory_records});
how completed navigation attempts update scene knowledge and episodic traces (\cref{sec:supp_memory_updates});
how the agent queries memory during online planning and post-attempt reflection (\cref{sec:supp_memory_queries});
and how memory empirically accumulates across multi-trajectory explorations (\cref{sec:supp_memory_example}).

\subsection{Record Layout and Evidence References}
\label{sec:supp_memory_records}

\begin{table}[t]
    \centering
    \footnotesize
    \setlength{\tabcolsep}{3pt}
    \caption{Unified memory schemas and stored record fields.}
    \label{tab:supp_memory_records}
    \begin{tabular}{@{}>{\raggedright\arraybackslash}m{0.27\columnwidth}m{\dimexpr0.73\columnwidth-6pt\relax}@{}}
        \toprule
        Track / Record & Stored fields and evidence references \\
        \midrule
        \multicolumn{2}{@{}l}{\textit{Observation Anchor (Immutable Evidence)}} \\[2pt]
        Observation & Scene-global ID, source attempt ID, local step, 6-DoF pose, camera intrinsics/extrinsics, and filesystem image references. \\
        \midrule
        \multicolumn{2}{@{}l}{\textit{Scene Knowledge ($\mathcal{S}$)}} \\[2pt]
        Place & Unique ID, parent region ID, member observation IDs, visual description, representative viewpoint image references. \\
        Region & Unique ID, semantic room category label, natural-language architectural summary, and member place IDs. \\
        Object & Instance ID, semantic category, 3D metric bounding box, fused point cloud, and multi-view RGB/mask evidence. \\
        Transition & Executed trajectory segment, connected place IDs, and verified metric traversal displacement. \\
        \midrule
        \multicolumn{2}{@{}l}{\textit{Episodic Experience ($\mathcal{H}$)}} \\[2pt]
        Decision Event & Event ID, task goal $g$, observation ID, chosen view/action, recorded rationale, execution feedback, and next observation ID. \\
        Reflection Record & Linked decision ID, diagnostic failure category, spatial correction brief, and referenced validation viewpoints. \\
        \midrule
        \multicolumn{2}{@{}l}{\textit{Associative Links ($\mathcal{L}$)}} \\[2pt]
        Associative Indexes & Bi-directional observation-to-event index, entity-to-event attribution index, and hierarchical region-membership lookup. \\
        \bottomrule
    \end{tabular}
\end{table}

A fundamental principle of RetroNav is that an agent's experience contains two complementary streams: objective evidence about the physical environment, and subjective interpretations made toward specific goals. \cref{tab:supp_memory_records} specifies the database schemas concretizing this separation across $\mathcal{S}$, $\mathcal{H}$, and $\mathcal{L}$.

\textbf{Stable observation anchors.}
Every perceptual input imported into memory is anchored by an immutable \emph{observation} record $o$, indexed by a scene-global integer ID, source attempt ID, local step, and 6-DoF sensor pose. Sensor payloads (RGB, depth, instance masks, and local point clouds) are retained as referenced filesystem assets, preventing data duplication across semantic records. Because raw sensor observations are immutable, downstream semantic interpretations can be iteratively revised without corrupting underlying empirical evidence.

\textbf{Physical and semantic scene knowledge ($\mathcal{S}$).}
Scene knowledge maintains the persistent spatial representation:
\begin{itemize}[leftmargin=*,itemsep=1pt,topsep=1pt]
    \item \emph{Places} group observations within shared local vantages, retaining 1--4 representative multi-view images and synthesizing view-specific spatial descriptions.
    \item \emph{Regions} cluster confirmed places into architectural enclosures (e.g., bedrooms, corridors), decoupling viewpoint vantages from observed room bounds.
    \item \emph{Objects} store 3D metric bounding boxes, fused point clouds, categories, and bounding-box image crops.
    \item \emph{Transitions} record verified traversability exclusively from physically executed displacements, guaranteeing collision-free path connectivity.
\end{itemize}

\textbf{Episodic decision traces ($\mathcal{H}$).}
Complementary to scene knowledge, episodic history preserves the temporal sequence of agent reasoning. Each decision event stores the task goal $g$, current observation ID, chosen view/action, explicit task interpretation and decision rationale, execution feedback, and next observation ID. When failure reflection is conducted (\cref{sec:reflection}), retrospective diagnoses and suggested corrections are appended as linked reflection records, preserving historical decision traces in their original context while augmenting them with diagnostic insights.

\textbf{Dual associative indexes ($\mathcal{L}$).}
The association layer connects $\mathcal{S}$ and $\mathcal{H}$ through two complementary links:
\begin{itemize}[leftmargin=*,itemsep=1pt,topsep=1pt]
    \item \emph{Provenance index}: links each decision event to its immediate observation, resolving its place and region membership. If topological places merge during map refinement, events automatically re-resolve their associations through their persistent observation anchors.
    \item \emph{Contextual attribution index}: distinguishes topological colocation (an event occurred at a place) from causal attribution (a decision explicitly cited a specific object or past experience), enabling selective retrieval during subsequent reasoning.
\end{itemize}

\subsection{Incremental Update Pipeline}
\label{sec:supp_memory_updates}

Unified memory is updated by extracting object evidence from RGB images, associating that evidence in 3D, organizing objects and observations into places and regions, and linking the resulting scene knowledge to navigation experience. The update unit is one completed navigation attempt, whether successful or unsuccessful. Each attempt contributes its saved observations, decisions, and actual movements to the previous committed memory. \cref{alg:supp_memory_update} summarizes the complete pipeline.

\begin{algorithm}[t]
    \caption{Incremental Unified Memory Update}
    \label{alg:supp_memory_update}
    \begin{algorithmic}[1]
        \Require Attempt log (four-view RGB-D, poses, decisions, and execution outcomes), prior memory $\mathcal{M}_{t-1}=(\mathcal{S}_{t-1},\mathcal{H}_{t-1},\mathcal{L}_{t-1})$, and ObjectMap $\mathcal{O}_{t-1}$.
        \Ensure Updated unified memory $\mathcal{M}_t$ and ObjectMap $\mathcal{O}_t$.
        \vspace{2pt}
        \State Assign stable observation IDs to the attempt's RGB-D observations and poses, retaining decision and movement records.
        \State Annotate salient objects in each RGB image to obtain categories, descriptions, and segmentation masks.
        \State Back-project valid masked depth into world coordinates, then downsample and denoise the point clouds to obtain the current trajectory's 3D object observations $\mathcal{W}_t$.
        \State Associate $\mathcal{W}_t$ with $\mathcal{O}_{t-1}$ through geometric overlap, fusing matched observations and adding unmatched instances to obtain $\mathcal{O}_t$.
        \State Invoke Memory Builder with the current trajectory, $\mathcal{O}_t$, and relevant context from $\mathcal{S}_{t-1}$ to produce $\mathcal{S}_t$.
        \State Organize logged decisions into an experience, preserving observation references, task interpretations, actions, and outcomes; append it to obtain $\mathcal{H}_t$.
        \State Update observation-to-event and entity-to-event indexes using scene evidence links and explicit decision references to obtain $\mathcal{L}_t$.
        \State Validate observation ownership and evidence references; save $\mathcal{M}_t=(\mathcal{S}_t,\mathcal{H}_t,\mathcal{L}_t)$ and $\mathcal{O}_t$.
        \State \Return $\mathcal{M}_t,\mathcal{O}_t$.
    \end{algorithmic}
\end{algorithm}

\textbf{Object annotation and segmentation.}
We first use GPT-5.6-Terra~\cite{openai2026sol} to propose up to four salient objects in each RGB image, providing their bounding boxes, categories, and appearance descriptions. 
We then use these boxes to prompt Segment Anything (SAM)~\cite{kirillov2023sam}, which refines the object extents into pixel-level segmentation masks. The resulting masks and their source observations are used for subsequent 3D object fusion.

\textbf{Depth projection and point-cloud filtering.}
We first filter SAM masks using a quality score threshold of 0.95 across all experiments. 
Valid depth pixels within each retained mask are back-projected into camera coordinates and then transformed into world coordinates using the camera calibration and robot pose. 
Each object point cloud is voxel-downsampled at 0.025\,m and clustered using DBSCAN. We retain the largest non-noise cluster and discard it if it contains fewer than 20 points.

\textbf{Cross-view association and object fusion.}
For each point cloud $\mathcal{W}_t^i$, we first select spatially nearby object candidates from $\mathcal{O}_{t-1}$. Its association score with a candidate $\mathcal{O}_{t-1}^k$ is defined as
\begin{equation}
\label{eq:supp_object_association}
s_{ik}=\frac{1}{|\mathcal{W}_t^i|}
\sum_{\mathbf{p}\in\mathcal{W}_t^i}
\mathbf{1}\!\left[
\min_{\mathbf{q}\in\mathcal{O}_{t-1}^k}
\|\mathbf{p}-\mathbf{q}\|_2\leq\delta
\right],
\end{equation}
where $\delta=0.05$\,m and $\mathbf{1}[\cdot]$ is the indicator function. We select the candidate with the highest association score. If its score exceeds 0.5, we fuse the two point clouds and combine their associated visual evidence. Otherwise, we add $\mathcal{W}_t^i$ as a new object instance. This process yields the updated object map $\mathcal{O}_t$.

\textbf{Scene knowledge update.}
Given $\mathcal{O}_t$, Memory Builder updates scene knowledge using the current trajectory and $\mathcal{S}_{t-1}$. Historical scene knowledge contains extensive image evidence and textual descriptions, much of which is unrelated to the current trajectory. 
To reduce token usage and processing time, we select relevant historical context before invoking the builder. Specifically, we retrieve places and objects within 1.5\,m of the current trajectory. For each object, we select a subset of historical observation images with distinct viewpoints.
The resulting input includes the current trajectory, retrieved place descriptions and their associated images, object records and their representative images, and the accumulated bird's-eye-view (BEV) map.

Using this context, Memory Builder assigns the current trajectory's observations to existing places or creates new places. 
It also merges nearby places judged to represent the same location. A place merge preserves all member observations and updates the consolidated place description. 
The builder then creates or updates regions based on spatial layout and semantic evidence, assigning places and objects to regions separately. Finally, the system adds traversability links supported by movements actually executed along the current trajectory, producing $\mathcal{S}_t$.

\textbf{Experience organization and scene association.}
We organize the current trajectory's log into a navigation experience containing the task goal, starting state, and temporally ordered decision events. Each event records its observation ID, task interpretation, selected view or action, explicit decision rationale, execution outcome, and subsequent observation. Historical judgments are preserved as originally recorded. 
Later reflections are appended as separate records referencing the relevant events and supporting evidence. The new experience is added to the existing collection to obtain $\mathcal{H}_t$.

Observation IDs serve as shared anchors between navigation experience and scene knowledge. We first build an observation-to-event index linking each observation to its corresponding decision events. 
We then construct reverse entity-to-event indexes through the observation's place membership, the place's region membership, and the object instances supported by that observation. These indexes allow a historical decision to retrieve related scene evidence and allow a place, region, or object to retrieve relevant decisions across navigation experiences. 
We distinguish two types of association: scene-context links established through shared observations, and explicit references to objects, images, or historical experiences recorded in a decision. The former describe the decision's scene context, while the latter identify evidence explicitly used to support it.

When places merge or region assignments change, the affected indexes are updated through the unchanged observation IDs. 
After object fusion, retained source evidence similarly links relevant events to the same object instance. 
Scene knowledge can therefore evolve while historical decisions and their original evidence remain accessible, yielding the updated association set $\mathcal{L}_t$.

\textbf{Integrity validation and commit.}
Before saving the updated memory, the system checks that every new observation belongs to exactly one place, region and object references exist, place merges are valid, and representative images and image-supported descriptions refer to observations belonging to the corresponding place. Event and association references must resolve to existing observations, entities, or historical events, and forward and reverse indexes must agree. Transitions are added from the actual execution log. The resulting scene knowledge, navigation experiences, and associations are saved together with the updated ObjectMap for subsequent navigation and reflection.

\begin{figure*}[!t]
    \centering
    \begin{subfigure}{\textwidth}
        \centering
        \includegraphics[width=\linewidth]{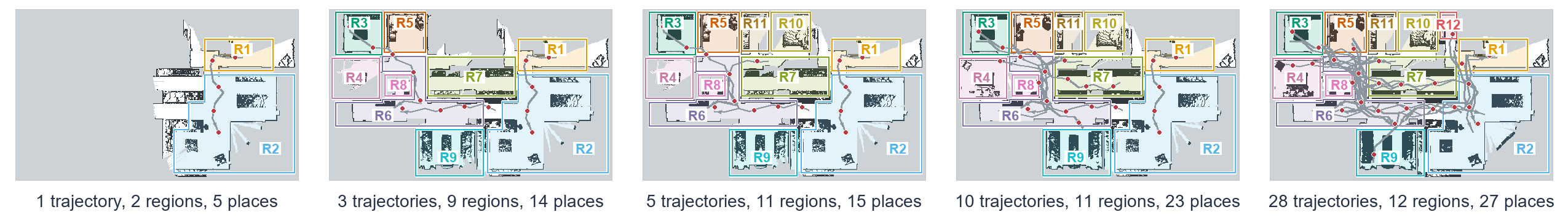}
        \caption{R2R scene \texttt{zsNo4HB9uLZ}.}
        \label{fig:supp_memory_growth_r2r}
    \end{subfigure}\par
    \vspace{5pt}
    \begin{subfigure}{\textwidth}
        \centering
        \includegraphics[width=\linewidth]{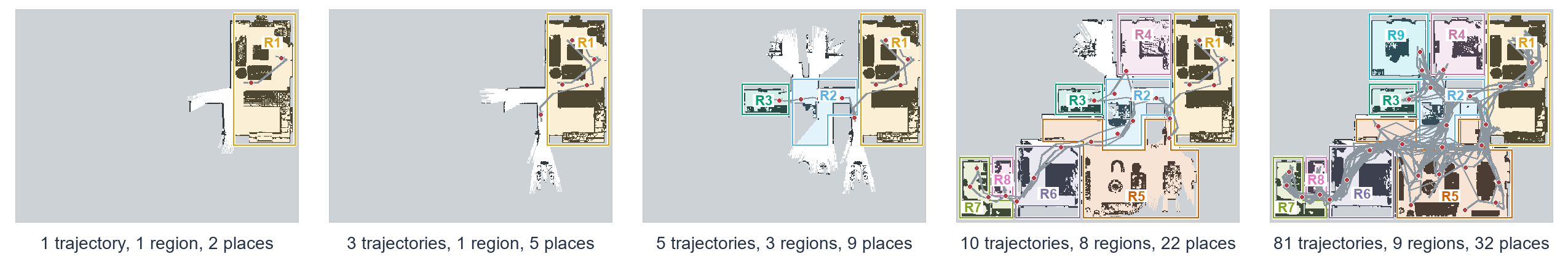}
        \caption{GOAT-Bench scene \texttt{5cdEh9F2hJL}.}
        \label{fig:supp_memory_growth_goat}
    \end{subfigure}
    \caption{Progressive scene-memory accumulation. Gray lines show navigation trajectories, red dots mark places, and colored polygons indicate regions.}
    \label{fig:supp_memory_growth}
\end{figure*}

\begin{figure*}[!t]
    \centering
    \begin{subfigure}[t]{0.325\textwidth}
        \centering
        \includegraphics[width=\linewidth]{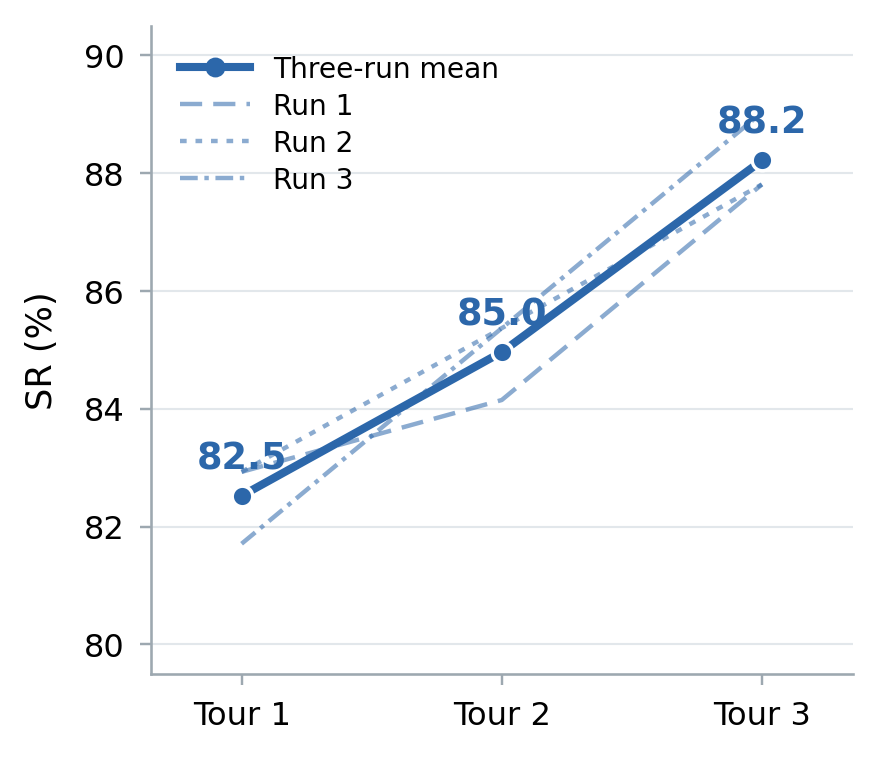}
        \caption{Final SR.}
        \label{fig:supp_memory_tours_sr}
    \end{subfigure}\hfill
    \begin{subfigure}[t]{0.325\textwidth}
        \centering
        \includegraphics[width=\linewidth]{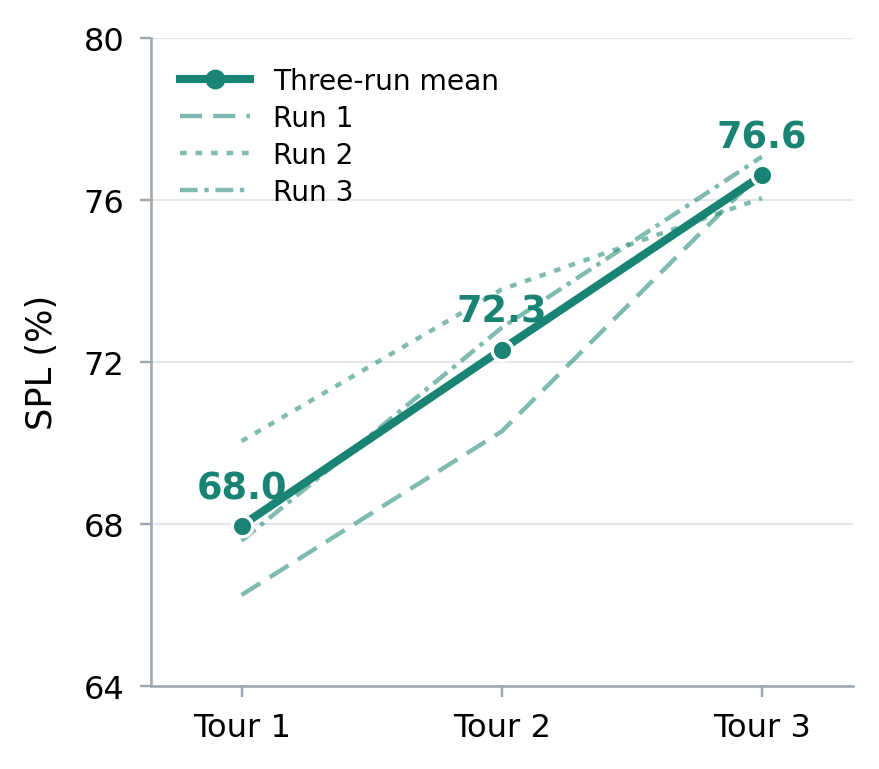}
        \caption{Final SPL.}
        \label{fig:supp_memory_tours_spl}
    \end{subfigure}\hfill
    \begin{subfigure}[t]{0.325\textwidth}
        \centering
        \includegraphics[width=\linewidth]{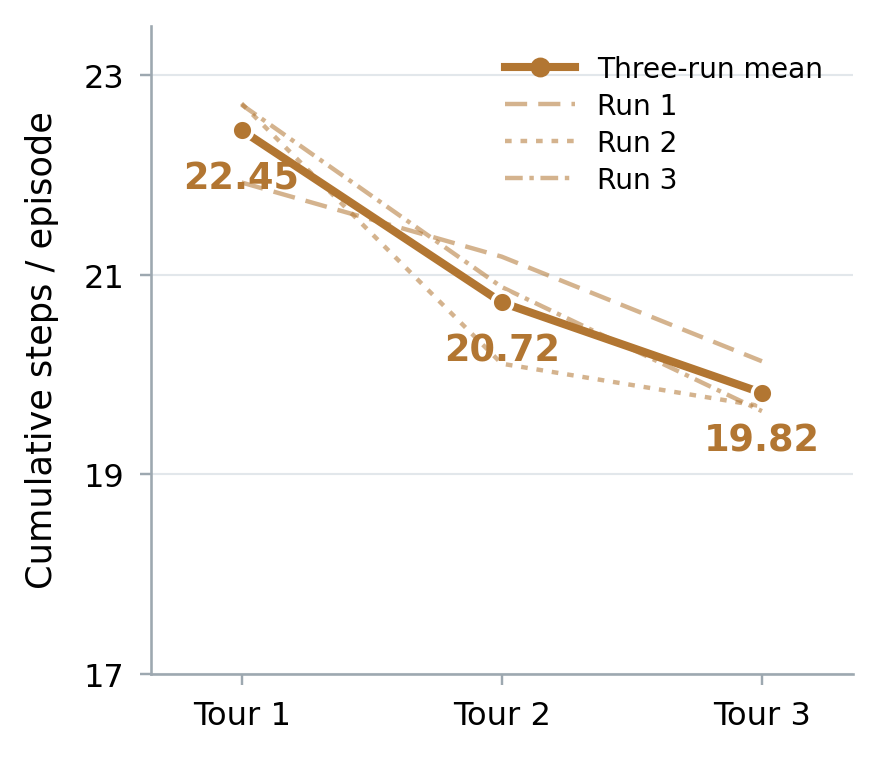}
        \caption{Navigation Steps.}
        \label{fig:supp_memory_tours_steps}
    \end{subfigure}
    \caption{Navigation performance over repeated tours in IR2R-CE. Solid lines show the mean across three independent runs; dashed lines show individual runs.}
    \label{fig:supp_memory_tours}
\end{figure*}

\begin{table*}[!t]
    \centering
    \footnotesize
    \setlength{\tabcolsep}{1.5pt}
    \caption{Retry performance and navigation cost per episode (subtask for GOAT-Bench). First and Final report the corresponding attempt; Cumulative includes all attempts. SR and SPL are percentages. PL denotes path length in meters. }
    \label{tab:supp_retry_actions}
    \label{tab:supp_cumulative_spl}
    \begin{tabular*}{\textwidth}{@{\extracolsep{\fill}}ll*{11}{c}@{}}
        \toprule
        \raisebox{-1.4ex}[0pt][0pt]{Benchmark} & \raisebox{-1.4ex}[0pt][0pt]{Model} & \multicolumn{4}{c}{First} & \multicolumn{4}{c}{Final} & \multicolumn{3}{c}{Cumulative} \\
        \cmidrule(lr){3-6}\cmidrule(lr){7-10}\cmidrule(lr){11-13}
        & & SR & SPL & Steps & PL (m) & SR & SPL & Steps & PL (m) & SPL & Steps & PL (m) \\
        \midrule
        \raisebox{-\normalbaselineskip}[0pt][0pt]{R2R-CE} & GPT-6-Astra & 77.3 & 71.5 & 12.1 & 9.7 & 84.0 & 75.4 & 16.2 & 13.5 & 66.3 & 17.6 & 14.3 \\
         & GPT-5.6-Sol & 66.3 & 52.2 & 14.3 & 11.4 & 76.3 & 66.4 & 18.3 & 15.0 & 52.0 & 21.2 & 17.1 \\
         & Qwen3.8-27B & 50.7 & 39.6 & 14.5 & 11.7 & 58.0 & 49.5 & 18.7 & 15.5 & 38.3 & 21.6 & 17.6 \\
        \midrule
        \raisebox{-\normalbaselineskip}[0pt][0pt]{RxR-CE} & GPT-6-Astra & 75.3 & 69.8 & 15.9 & 13.6 & 84.0 & 70.9 & 16.3 & 14.2 & 66.9 & 21.2 & 18.2 \\
         & GPT-5.6-Sol & 62.0 & 55.5 & 18.0 & 15.2 & 76.7 & 61.0 & 17.8 & 15.1 & 54.2 & 26.1 & 22.0 \\
         & Qwen3.8-27B & 56.7 & 40.2 & 19.4 & 16.9 & 64.0 & 46.4 & 19.9 & 17.1 & 35.9 & 29.4 & 25.5 \\
        \midrule
        \raisebox{-\normalbaselineskip}[0pt][0pt]{HM3D-OVON} & GPT-6-Astra & 61.8 & 40.3 & 8.4 & 7.0 & 66.9 & 50.2 & 9.6 & 8.2 & 40.5 & 10.3 & 8.6 \\
         & GPT-5.6-Sol & 56.7 & 45.0 & 8.3 & 6.9 & 60.1 & 44.5 & 9.9 & 8.3 & 41.8 & 10.7 & 8.9 \\
         & Qwen3.8-27B & 51.1 & 28.1 & 8.9 & 7.6 & 55.4 & 45.8 & 10.1 & 8.5 & 28.5 & 11.2 & 9.5 \\
        \midrule
        \raisebox{-\normalbaselineskip}[0pt][0pt]{IR2R-CE} & GPT-6-Astra & 80.0 & 67.7 & 14.8 & 12.0 & 89.2 & 74.5 & 15.0 & 12.2 & 68.5 & 18.6 & 15.1 \\
         & GPT-5.6-Sol & 72.4 & 64.2 & 15.4 & 12.5 & 84.1 & 71.9 & 15.5 & 12.5 & 65.6 & 20.3 & 16.5 \\
         & Qwen3.8-27B & 50.1 & 39.9 & 16.8 & 13.1 & 57.2 & 48.1 & 17.2 & 13.3 & 36.0 & 26.3 & 20.5 \\
        \midrule
        \raisebox{-\normalbaselineskip}[0pt][0pt]{GOAT-Bench} & GPT-6-Astra & 81.9 & 57.2 & 8.4 & 7.7 & 84.1 & 58.4 & 9.1 & 8.5 & 55.2 & 10.0 & 9.2 \\
         & GPT-5.6-Sol & 79.0 & 55.8 & 8.6 & 7.6 & 82.3 & 57.6 & 9.0 & 8.0 & 53.3 & 10.6 & 9.4 \\
         & Qwen3.8-27B & 70.1 & 43.6 & 9.2 & 8.2 & 74.4 & 45.7 & 9.5 & 8.4 & 41.0 & 11.8 & 10.5 \\
        \bottomrule
    \end{tabular*}
\end{table*}

\begin{table*}[!t]
    \centering
    \footnotesize
    \setlength{\tabcolsep}{2pt}
    \caption{Total inference cost per episode (subtask for GOAT-Bench) across all attempts. / means the component was not used.}
    \label{tab:supp_retry_inference}
    \begin{tabular*}{\textwidth}{@{\extracolsep{\fill}}lc*{9}{c}@{}}
        \toprule
        \raisebox{-1.4ex}[0pt][0pt]{Benchmark} & \raisebox{-1.4ex}[0pt][0pt]{Model} & \raisebox{-1.4ex}[0pt][0pt]{Calls} & \multicolumn{2}{c}{Tokens (K)} & \multicolumn{6}{c}{Cumulative model-call time (s)} \\
        \cmidrule(lr){4-5}\cmidrule(lr){6-11}
        & & & Input & Output & Navigation & Retry brief & Evaluator & Object annot. & Memory builder & Total \\
        \midrule
        \raisebox{-0.5\normalbaselineskip}[0pt][0pt]{R2R-CE} & GPT-5.6-Sol & 65.3 & 663.0 & 12.2 & 302.0 & 47.7 & 49.9 & / & / & 399.6 \\
         & GPT-6-Astra & 44.8 & 445.0 & 5.8  & 218.0 & 19.5 & 41.1 & /     & /     & 278.6 \\
        \midrule
        \raisebox{-0.5\normalbaselineskip}[0pt][0pt]{RxR-CE} & GPT-5.6-Sol & 53.0 & 629.5 & 9.2 & 366.5 & 25.0 & 60.9 & / & / & 452.4 \\
         & GPT-6-Astra & 39.3 & 382.4 & 3.6 & 247.8 & 5.8 & 57.2 & / & / & 310.8 \\
        \midrule
        \raisebox{-0.5\normalbaselineskip}[0pt][0pt]{HM3D-OVON} & GPT-5.6-Sol & 29.2 & 199.1 & 5.3 & 162.0 & 23.1 & 27.6 & / & / & 212.7 \\
         & GPT-6-Astra & 26.7 & 164.9 & 3.1 & 180.9 & 26.2 & 30.2 & / & / & 237.3 \\
        \midrule
        \raisebox{-0.5\normalbaselineskip}[0pt][0pt]{IR2R-CE} & GPT-5.6-Sol & 56.5 & 902.6 & 26.7 & 393.7 & 27.6 & 55.7 & 484.6 & 140.0 & 1101.6 \\
         & GPT-6-Astra & 50.2 & 860.0 & 19.2 & 308.0 & 18.5 & 47.1 & 482.0 & 131.5 & 987.1 \\
        \midrule
        \raisebox{-0.5\normalbaselineskip}[0pt][0pt]{GOAT-Bench} & GPT-5.6-Sol & 36.6 & 670.8 & 17.6 & 197.3 & 13.6 & 30.3 & 263.7 & 66.7 & 571.6 \\
         & GPT-6-Astra & 36.4 & 745.8 & 14.5 & 156.3 & 11.6 & 28.0 & 271.1 & 62.5 & 529.5 \\
        \bottomrule
    \end{tabular*}
\end{table*}

\subsection{Query Interfaces and Context Budgets}
\label{sec:supp_memory_queries}

During deployment, the agent queries unified memory for two distinct cognitive purposes: online spatial grounding during forward navigation (\cref{sec:navigation}), and causal mistake diagnosis during post-attempt reflection (\cref{sec:reflection}).

\textbf{Online spatial context.}
To support forward viewpoint selection without exceeding prompt limits, the system dynamically maintains a local context window encompassing places within a horizontal radius of 3.0\,m and vertical threshold of 1.5\,m (capped at 8 places). A 0.5\,m hysteresis margin prevents rapid context oscillations across topological boundaries. Within this local window, objects are grouped into at most 20 semantic categories. If local information is insufficient to resolve movement direction, the View Selection Agent invokes \texttt{search\_memory} (querying 1--4 phrases, returning up to 6 ranked entity summaries via lexical and exact-token matching) and \texttt{read\_memory} (fetching detailed descriptions and a representative high-resolution RGB image for up to 4 selected Place/Object IDs).

\textbf{Cross-task experience retrieval.}
When facing ambiguous intersections or misleading visual landmarks, the agent queries the associative layer $\mathcal{L}$ using candidate spatial entities and current subgoals as anchors. The retrieval engine identifies relevant historical decision events located at the same place or citing the same entities. The interface returns up to three concise event summaries and extracts up to two detailed decision fragments, providing the historical goal, action choice, execution consequence, and linked reflection without congesting context windows.

\textbf{Reflection query protocol.}
During post-attempt reflection, the Retrospection Agent operates under distinct query bounds tailored for forensic diagnosis: \texttt{search\_memory} returns up to 5 comprehensive records, while \texttt{read\_images} allows inspection of historical and newly committed multi-view evidence.
Crucially, there is a clear operational boundary: online navigation queries only access memory committed prior to the active attempt, preventing unvalidated within-attempt leaks, whereas reflection accesses both prior history and newly committed evidence from the failed run to construct the retry brief.

\subsection{Spatial Expansion and Place-Level Enrichment}
\label{sec:supp_memory_example}

\cref{fig:supp_memory_growth} provides empirical evidence of how completed trajectories progressively construct a coherent spatial representation. Rather than growing uniformly, memory accumulation manifests in two complementary structural dimensions: \emph{spatial expansion} into previously unvisited areas, and \emph{place-level enrichment} within existing regions.

\textbf{Spatial expansion across unseen layouts.}
In the R2R-CE scene (\cref{fig:supp_memory_growth_r2r}), the initial trajectory establishes only 2 regions and 5 places. As exploration progresses, the agent extends its topological frontier across corridors and doorways, reaching 12 regions and 27 places after 28 trajectories. The early visits rapidly discover major architectural rooms, expanding physical coverage.

\textbf{Place-level enrichment within confirmed regions.}
Concurrently, subsequent trajectories enrich the internal topological granularity of existing regions without triggering spurious region proliferation. For example, between trajectories 5 and 10 in R2R-CE, the total region count remains strictly stable at 11, while the place count increases from 15 to 23. The agent grounds multiple distinct vantage points within known rooms, providing denser viewpoints for landmark verification. The GOAT-Bench sequence (\cref{fig:supp_memory_growth_goat}) exhibits identical dual-stage dynamics: initial attempts establish 5 places within a single region, before subsequent attempts expand the map to 9 regions and 32 places. Stationary or failed attempts contribute observation evidence to semantic instances without introducing redundant trajectory edges, demonstrating robust structural stability.

\section{Additional Experimental Results}
\label{sec:supp_additional_results}

\subsection{Navigation Benefits over Repeated Tours}
\label{sec:supp_memory_tours}

To verify whether cumulative memory translates into tangible navigation advantages when revisiting familiar environments, \cref{fig:supp_memory_tours} evaluates multi-tour navigation on scene \texttt{QUCTc6BB5sX} from the IR2R-CE benchmark. To examine long-horizon stability under extended scene persistence without resets, we evaluate multi-tour navigation across three independent runs with GPT-5.6-Sol, where each run traverses three consecutive tours of the same 82 reference routes under varied linguistic instructions. Memory is accumulated continuously across tours without resets.

Across the three tours, final Success Rate (SR) steadily climbs from 82.5\% to 88.2\% (+5.7 percentage points), and final SPL rises from 68.0\% to 76.6\% (+8.6 percentage points). 
Crucially, this performance improvement is accompanied by a significant reduction in navigation cost: mean cumulative navigation steps per episode decrease by 11.7\%. Because the agent reuses confirmed landmarks and avoids previously explored dead ends, it navigates more decisively, achieving higher task completion while taking more direct paths.

\subsection{Multi-Attempt Recovery and Efficiency}
\label{sec:supp_retry_cost}
\label{sec:supp_cumulative_spl}

\textbf{Multi-attempt recovery gains.}
\cref{tab:supp_retry_actions} separates first-attempt and final-attempt performance from cumulative navigation cost across all benchmarks. Reflective retry consistently raises task completion by allowing the agent to diagnose past failure modes and adjust its navigation strategy. 
On IR2R-CE, reflective retry raises GPT-6-Astra SR from 80.0\% to 89.2\% and GPT-5.6-Sol from 72.4\% to 84.1\%. On R2R-CE, GPT-6-Astra increases SR from 77.3\% to 84.0\%, while on HM3D-OVON, it raises SR from 61.8\% to 66.9\%. 
Across all benchmarks, the agent actively leverages retrospective diagnoses to correct misjudgments rather than repeating identical trajectory errors.

\textbf{Path efficiency and movement cost.}
To account for travel during failed attempts, we extend the path denominator of SPL to include every attempt at the same task:
{
\begin{equation}
    \mathrm{Cumulative\ SPL}
    = \frac{1}{N}\sum_{i=1}^{N}
    \frac{S_i d_i}{\max\left(d_i,\sum_{k=1}^{K_i}\ell_{ik}\right)},
    \label{eq:cumulative_spl}
\end{equation}
}
where $S_i$ denotes success on the final attempt, $d_i$ is the shortest navigable distance from the task's starting pose to its goal, $K_i$ is the number of attempts, and $\ell_{ik}$ is the traveled distance in attempt $k$. 
Tasks without a retry retain their standard SPL. The simulator resets the agent to the starting pose between attempts; this reset does not add a traveled path segment. 
As reported in \cref{tab:supp_retry_actions}, Cumulative SPL penalizes total distance traveled during prior unsuccessful attempts while rewarding decisive recovery. Across all benchmarks, RetroNav maintains strong cumulative efficiency despite multi-attempt exploration. For example, cumulative SPL reaches 68.5\% on IR2R-CE with GPT-6-Astra and 65.6\% with GPT-5.6-Sol, with 18.6 and 20.3 cumulative steps per task. 
On GOAT-Bench, GPT-6-Astra achieves 55.2\% cumulative SPL compared with 58.4\% final-attempt SPL, demonstrating that retries resolve tasks through targeted recovery rather than unguided wandering.

\textbf{Inference cost.}
\cref{tab:supp_retry_inference} reports mean calls, token counts, and cumulative model-call time per episode (subtask for GOAT-Bench), including all attempts and tool rounds. 
For time, we first sum model-call durations within each task and then average over complete task records. Component times use the same records, with zero cost when a component is not invoked. 
Inference breakdown shows that background object annotation remains decoupled and efficient, handled consistently by GPT-5.6-Terra at approximately 482\,s per episode on IR2R-CE. 
Because GPT-6-Astra exhibits higher first-attempt accuracy and reduced exploration detours, it requires significantly fewer model calls and tokens, reducing cumulative navigation latency across persistent sessions.

\subsection{Joint View and Waypoint Selection}
\label{sec:supp_joint_selection}

\balance

We evaluate joint view and waypoint selection with Qwen on R2R-CE. The two-stage scheme first selects a view from four RGB images and then selects a waypoint within that view (\cref{sec:navigation}). The joint scheme receives the four images and their waypoint candidates, and directly outputs both a view ID and a waypoint ID, removing one model call per decision.
We report SR, SPL, OSR, model calls, tokens, and cumulative model-call time averaged per episode; robot execution time is excluded. \cref{tab:supp_joint_selection} compares the two schemes.

\begin{table}[t]
    \centering
    \footnotesize
    \setlength{\tabcolsep}{2pt}
    \caption{Two-stage versus joint view and waypoint selection with Qwen on R2R-CE. Costs are averaged per episode. Tokens include input and output; time is cumulative model-call time. -- denotes an unreported value.}
    \label{tab:supp_joint_selection}
    \begin{tabular*}{\columnwidth}{@{\extracolsep{\fill}}lcccccc@{}}
        \toprule
        Scheme & SR $\uparrow$ & SPL $\uparrow$ & OSR $\uparrow$ & Steps $\downarrow$ & Tokens (K) $\downarrow$ & Time (s) $\downarrow$ \\
        \midrule
        Two-stage & \textbf{56.0} & \textbf{48.77} & \textbf{72.0} & \textbf{17.65} & \textbf{265.41} & 156.77 \\
        Joint & 47.0 & 39.65 & 61.0 & 18.29 & 390.92 & \textbf{125.93} \\
        \bottomrule
    \end{tabular*}
\end{table}

As shown in \cref{tab:supp_joint_selection}, the joint scheme performs worse than the two-stage scheme across the navigation metrics. It requires a longer trajectory with approximately one additional decision step per episode. 
Although joint selection reduces overall latency by eliminating one round-trip API call per step, this latency reduction comes at the expense of substantially higher cumulative token consumption (390.92\,K vs.\ 265.41\,K).

The main reason is that the reported token count is accumulated over all decision steps, including both input and output tokens. As navigation proceeds, the input context becomes longer because it contains an expanding history of observations, actions, and execution feedback. 
Therefore, an additional step near the end of an episode carries a larger token cost than an early step. Even though the joint scheme uses only about one more step on average, this extra step introduces a complete additional context window and substantially increases the cumulative input tokens. 
Consequently, the joint scheme consumes more tokens overall despite using fewer model calls per decision. These results suggest that separating view selection and waypoint selection provides more reliable navigation decisions and better controls the accumulation of context over long trajectories.

\end{document}